\documentclass[]{bytedance_seed}

\usepackage{hyperref}
\usepackage{cleveref}
\usepackage{verbatim}

\usepackage{wrapfig}  
\usepackage{graphicx}
\usepackage{floatrow}
\usepackage{subcaption}
\usepackage{listings}
\usepackage{algorithm}

\usepackage{subcaption} %

\usepackage[toc,page,header]{appendix}

\usepackage{minitoc}

\newcommand{\ourmodel}{BAGEL}

\title{Emerging Properties in Unified Multimodal Pretraining}

\author{%
\parbox{\textwidth}{\centering
Chaorui Deng$^{*1}$, Deyao Zhu$^{*1}$, Kunchang Li$^{*2\ddagger}$, Chenhui Gou$^{*3\ddagger}$, Feng Li$^{*4\ddagger}$\\[2mm]
Zeyu Wang$^{5\ddagger}$, Shu Zhong$^{1}$, Weihao Yu$^{1}$,Xiaonan Nie$^{1}$, Ziang Song$^{1}$, Guang Shi$^{1\S}$\\[2mm]Haoqi Fan$^{*\dagger}$
}}

\affiliation{%
\parbox{\textwidth}{\centering\small
$^1$ByteDance Seed,
$^2$Shenzhen Institutes of Advanced Technology,
$^3$Monash University\\[1mm]
$^4$Hong Kong University of Science and Technology,
$^5$UC Santa Cruz
}}

\contribution[*]{Equal contribution}
\contribution[\S]{Corresponding Author}
\contribution[\dagger]{Project lead}

\footnotetext[3]{Work was done during their internship.}

\abstract{
Unifying multimodal understanding and generation has shown impressive capabilities in cutting‑edge proprietary systems. In this work, we introduce BAGEL, an open-source foundational model that natively supports multimodal understanding and generation. BAGEL is a unified, decoder-only model pretrained on trillions of tokens curated from large-scale interleaved text, image, video, and web data. When scaled with such diverse multimodal interleaved data, BAGEL exhibits emerging capabilities in complex multimodal reasoning. As a result, it significantly outperforms open-source unified models in both multimodal generation and understanding across standard benchmarks, while exhibiting advanced multimodal reasoning abilities such as free-form image manipulation, future frame prediction, 3D manipulation, and world navigation. In the hope of facilitating further opportunities for multimodal research, we share the key findings, pretraining details, data creation protocal, and release our code and checkpoints to the community.
}
\date{\today}
\checkdata[Corresponding]{\url{shiguang.sg@bytedance.com}}
\checkdata[Project Page]{\url{https://bagel-ai.org/}}

\begin{document}
\maketitle

\section{Introduction}
\label{sec:intro}

The field of unified multimodal understanding and generation has witnessed a surge in interest, with numerous research projects~\cite{sun2023emu, pan2025transfer, tong2024metamorph, chameleon, show-o, emu3, janus2024, shi2024llamafusion} demonstrate promising results in jointly optimizing generation and understanding benchmarks with a crafted unified architecture. 
While several efforts \cite{aghajanyan2023scaling, januspro2025, chameleon} attempt to scale up their unified models, they are still trained predominantly on image-text paired data from standard image generation and understanding tasks. 
Recent research~\cite{chen2025empirical} has revealed a substantial gap in unified multimodal understanding and generation between academic models and proprietary systems such as GPT-4o and Gemini 2.0, whose underlying techniques remain undisclosed. We argue that the key to close this gap lies in scaling with carefully structured multimodal interleaved data - integrates texts, images, videos and web sources. 
Our experiments reveal emerging properties as the interleaved multimodal pretraining scales up. Beyond enhancing core multimodal understanding and generation capabilities, the scaling also facilitates complex compositional abilities such as free-form visual manipulation and multimodal generation with long-context reasoning, paving the way for a broad spectrum of advanced functions.

\begin{figure}[H]     
  \vspace*{-0.05\textwidth}
  \hspace*{-0.145\textwidth}
  \includegraphics[width=1.28\textwidth]{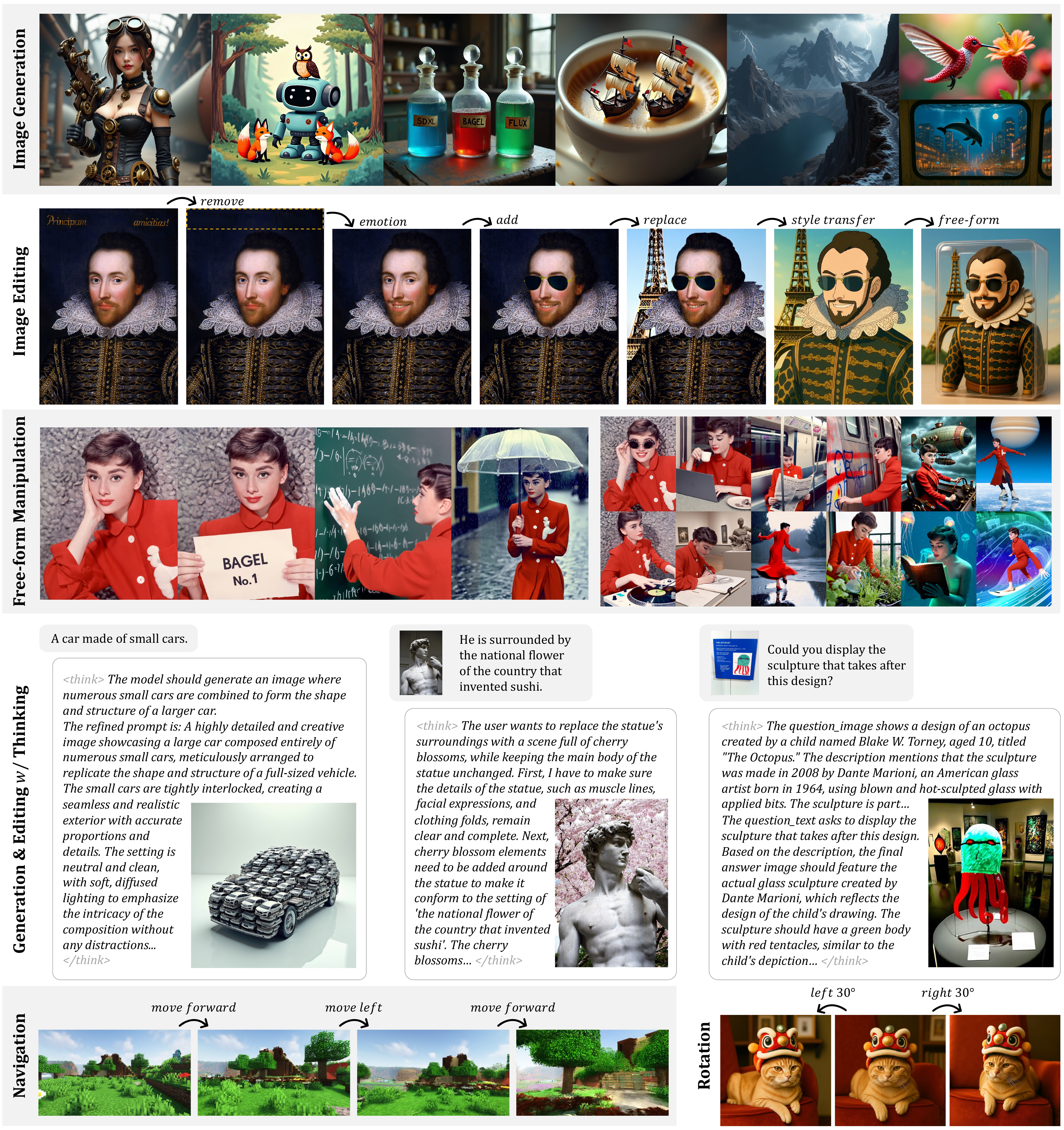}
  \caption{Showcase of the versatile abilities of the \textbf{BAGEL} model.}
  \label{fig:teaser}
\end{figure}

To realize this vision, we established a new protocol for scalable data sourcing, filtering, and construction of high-quality multimodal interleaved data. 
In addition to the web source, we incorporate video data that naturally provides pixel-level, conceptual, temporal, and physical continuity, which offers exclusive signals essential for acquiring grounded world knowledge at scale. Moreover, our interleaved format inherently includes tasks such as multimodal conversation, text-to-image/video, and image manipulation, enabling seamless integration of diverse generative data. 
Inspired by DeepSeek-R1~\cite{guo2025r1}, we further enrich the interleaved data with reasoning-oriented content to facilitate multi-modal reasoning, which enables seamless knowledge transfer between understanding and generation processes. As a result, the curated data captures rich world knowledge and nuanced cross-modal interaction content, equipping models with foundational capabilities in in-context prediction, world modeling, and complex multimodal reasoning.

Regarding architecture design, our primary objective is to maximize the capacity of the model without introducing heuristic bottlenecks or task-specific constraints commonly employed in previous models. Following this design philosophy, we adopt a Mixture-of-Transformer-Experts (MoT) architecture that employs selective activation of modality-specific parameters. Unlike some prior approaches~\cite{tong2024metamorph,pan2025transfer,emu2,dongdreamllm} that introduce bottleneck connectors between generation and understanding modules, our design enables long-context interaction between multimodal understanding and generation through shared self-attention operations. 
This bottleneck-free design enables effective scaling of training data and steps, allowing the model’s full capacity signals to emerge without being hindered or obscured by architectural constraints.

We present the Scala\underline{b}le Gener\underline{a}tive Co\underline{g}nitive Mod\underline{el} (\textbf{BAGEL}), an open‑source multimodal foundation model with 7B active parameters (14B total) trained on large‑scale interleaved multimodal data. BAGEL outperforms the current top‑tier open‑source VLMs~\cite{qwen2.5-vl,internvl2.5} on standard multimodal‑understanding leaderboards, and delivers text‑to‑image quality that is competitive with leading public generators such as SD3~\cite{SD3} and FLUX.1-dev~\cite{flux}.
Moreover, BAGEL demonstrates consistently superior qualitative results in classical image‑editing scenarios than the leading open-source models. More importantly, it extends to free-form visual manipulation, multiview synthesis, and world navigation, capabilities that constitute "world-modeling" tasks beyond the scope of previous image-editing models.
We showcase the qualitative performance in \Cref{fig:teaser}.

As BAGEL scales with interleaved multimodal pre‑training, we observe a clear emerging pattern: basic multimodal understanding and high‑fidelity generation converge first; next, complex editing and free-form visual manipulation abilities surface; finally, long-context reasoning starts to benefit multimodal understanding and generation, suggesting that previously independent atomic skills synergize into compositional reasoning across modalities. 
These emerging capabilities are not only supported by public benchmarks but are more distinctly revealed in our proposed IntelligentBench, and further verified by qualitative observations. 
These observations highlight that, while the optimization landscapes for understanding and generation remain partially decoupled, they can be jointly explored via shared self-attention context within a single transformer model, yielding a rich spectrum of capabilities in an open‑source system.
\vspace{5pt}
\section{Model}
\begin{figure*}[htb]
\centering
\includegraphics[width=0.9\textwidth]{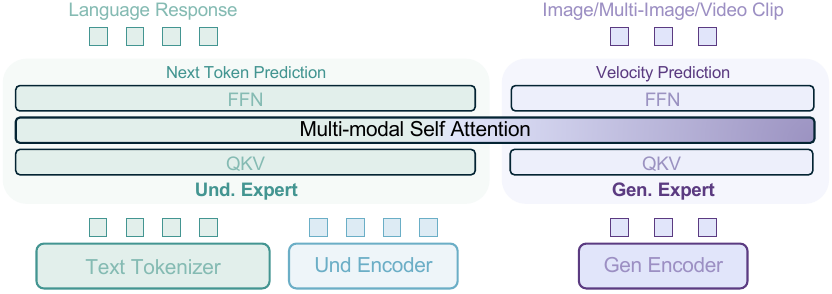}
\caption{
We use \textbf{two Transformer experts} to process understanding and generation information, and all tokens do shared multi-modal self attention in each Transformer block. We adopt two distinct encoders to separately capture semantic content and low-level pixel information for image understanding and generation tasks.
\vspace{-2pt}
} 
\label{fig:overview}
\end{figure*}

\vspace{5pt}
As illustrated in \Cref{fig:overview}, {BAGEL} adopts a MoT architecture comprising two transformer experts—one dedicated to multimodal understanding and the other to multimodal generation. Accordingly, the model employs two separate visual encoders: an understanding‑oriented encoder and a generation‑oriented encoder.
The two transformer experts operates on the same token sequence through the shared self-attention operation at every layer.
When predicting text tokens, {BAGEL} follows the Next‑Token‑Prediction paradigm, adhering to the well-established strengths of autoregressive language models. For visual token prediction, {BAGEL} adopts the {Rectified Flow}~\cite{liuflow,SD3,lipman2022flow} method following the best practice in the field of visual generation.
In the remainder of this section, we share the insights and motivations that shaped these design choices.

\vspace{5pt}
\subsection{Model Design Space}
\vspace{5pt}
Typical design choices for unified multi-modal generation and understanding models include:

\textbf{Quantized AR.} Autoregressive visual generation~\cite{janus2024,januspro2025,emu3,lu2024unified,vila-u,wu2024liquid,chameleon,xie2024muse,qu2024tokenflow} with discrete visual tokenizers~\cite{mentzer2023finite, huh2023improvedvqste,lee2022autoregressive,yu2022vectorquantized}. This line of methods leverage the Next-Token-Prediction paradigm for both text and visual token generation, 
which is straightforward to implement as it can directly utilize existing LLM infrastructures. 
Unfortunately, the visual generation quality of autoregressive models is empirically inferior to diffusion-based models. Furthermore, inference latency suffers due to the sequential nature of the autoregressive approach.

\textbf{External Diffuser.} LLM backbone combined with an external diffusion module~\cite{emu2,seed-x,dongdreamllm,pan2025transfer,tong2024metamorph}. This design connects pre-trained LLMs/VLMs to diffusion models via lightweight, trainable adapters. Typically, the language backbone autoregressively generates a set of latent tokens as "semantic condition" signals, which are then employed by the diffusion module to generate images. This setup often exhibits rapid convergence with minimal data consumption and may also yield competitive performance~\cite{pan2025transfer} on established benchmarks for multi-modal generation and understanding. Its primary drawback, however, is the compression of the LLM context into a relatively small number of latent tokens. 
This introduces an explicit bottleneck between understanding and generation modules, risking substantial information loss—particularly in long-context multimodal reasoning. 
Such a constraint might contradict the scaling philosophy of large foundational models.

\textbf{Integrated Transformer.} Unified integration of LLM and diffusion models within a single transformer~\cite{transfusion,janusflow2024,shi2024llamafusion,mot}. Driven by the complementary strengths of autoregressive transformers (powerful understanding/reasoning ability) and diffusion transformers (strong visual generation ability), this approach uses their common model architecture to enable seamless switching between both paradigms. 
Compared to the {External Diffuser} solution, it demands substantially higher training compute. Nonetheless, it offers a significant advantage by maintaining a bottleneck-free context throughout all transformer blocks, thereby enabling lossless interaction between the generation and understanding modules and is more amenable to scaling.

In this work, we argue that unified models have the capacity to learn richer multi-modal capabilities from large-scale interleaved multi-modal data—emergent abilities that are not captured by traditional benchmarks.
To this end, we choose the bottleneck-free \textit{Integrated Transformer} solution, which we believe to have greater potential in large-scale training settings and may better serve as the foundation model for long-context multimodel reasoning as well as reinforcement learning.

\subsection{Architecture}\label{sec:2.2}
Our backbone model is inherited from an LLM with a decoder-only transformer architecture. We choose Qwen2.5 LLM~\cite{qwen2.5} as the initialization for its superior performance~\cite{gandhi2025cognitive} and public availability. It adopts RMSNorm~\cite{zhang2019root} for normalization, SwiGLU~\cite{shazeer2020glu} for activation, RoPE~\cite{su2024roformer} for positional encoding, and GQA~\cite{ainslie2023gqa} for KV cache reduction. Moreover, we add the QK-Norm~\cite{dehghani2023scaling} in each attention block following the common practice in image/video generation models~\cite{SD3,flux,seawead2025seaweed}, which is effective in stabilizing the training process.

The visual information is represented from two aspects: 
\begin{itemize}

\item For \textit{visual understanding}, we leverage a ViT encoder to convert the raw pixels into tokens. We adopt SigLIP2-so400m/14~\cite{tschannen2025siglip} with a fixed $384$-resolution as the initialization of the ViT encoder. 
Building upon this, we first interpolate the position embedding and set $980\times 980$ as the maximum input size, and then integrate NaViT~\cite{dehghani2023patch} to enable processing of images at their native aspect ratios. A two-layer MLP connector is adopted to match the feature dimension of the ViT tokens and the LLM hidden states.

\item For \textit{visual generation}, we use a pre-trained VAE model from FLUX~\cite{flux} to convert images from pixel space to latent space and vice versa. The latent representation has a downsample ration of 8 and a latent channel of 16, and is then processed by a $2\times 2$ patch embedding layer to reduce the spatial size and match the hidden dimension of the LLM backbone. The VAE model is frozen during training.
\end{itemize}

Our framework applies 2D positional encoding to both ViT and VAE tokens prior to their integration into the LLM backbone.
For diffusion timestep encoding, we follow~\cite{causalfusion} and add a timestep embedding directly to the initial hidden states of VAE tokens, instead of using AdaLN as in conventional diffusion transformers~\cite{dit,SD3,flux}. This modification preserves performance while yielding a cleaner architecture.
Within the LLM, the text, ViT, and VAE tokens from understanding and generation tasks are interleaved according to the modality structure of input. 
For tokens belonging to the same sample, we employ a generalized version of the causal attention mechanism. These tokens are first partitioned into multiple consecutive splits, each containing tokens from a single modality (e.g., either text, ViT, or VAE). Tokens in one split may attend to all tokens in preceding splits. Inside each split, we adopt causal attention on text tokens, and keep the bidirectional attention on vision tokens.

\subsection{Generalized Causal Attention}

During training, an interleaved multimodal generation sample may contain multiple images.  
For each image, we prepare three sets of visual tokens:  
\begin{itemize}

\item \textbf{Noised VAE tokens}: VAE latents corrupted with diffusion noise, used exclusively for Rectified‑Flow training; the MSE loss is computed on this set.  

\item \textbf{Clean VAE tokens}: the original (noise‑free) latents, which serve as conditioning when generating subsequent image or text tokens.  

\item \textbf{ViT tokens}: obtained from the SigLIP2 encoder, which help to unify the input format across interleaved generation and understanding data and, empirically, to boost interleaved‑generation quality.  

\end{itemize}
For interleaved image or text generation, subsequent image or text tokens may attend to the clean VAE tokens and ViT tokens of preceding images, but \textit{not} to their noised VAE counterparts.  

For interleaved multi-image generation, we adopt the diffusion forcing strategy~\cite{chen2024diffusion}, which adds independent noise levels to different images and conditions each image on noisy representations of preceding images. Additionally, to enhance generation consistency, we randomly group consecutive images following~\cite{causalfusion} and apply full attention within each group. The noise level is the same inside each group.

\begin{figure}[ht]
\centering
\includegraphics[width=1.0\textwidth]{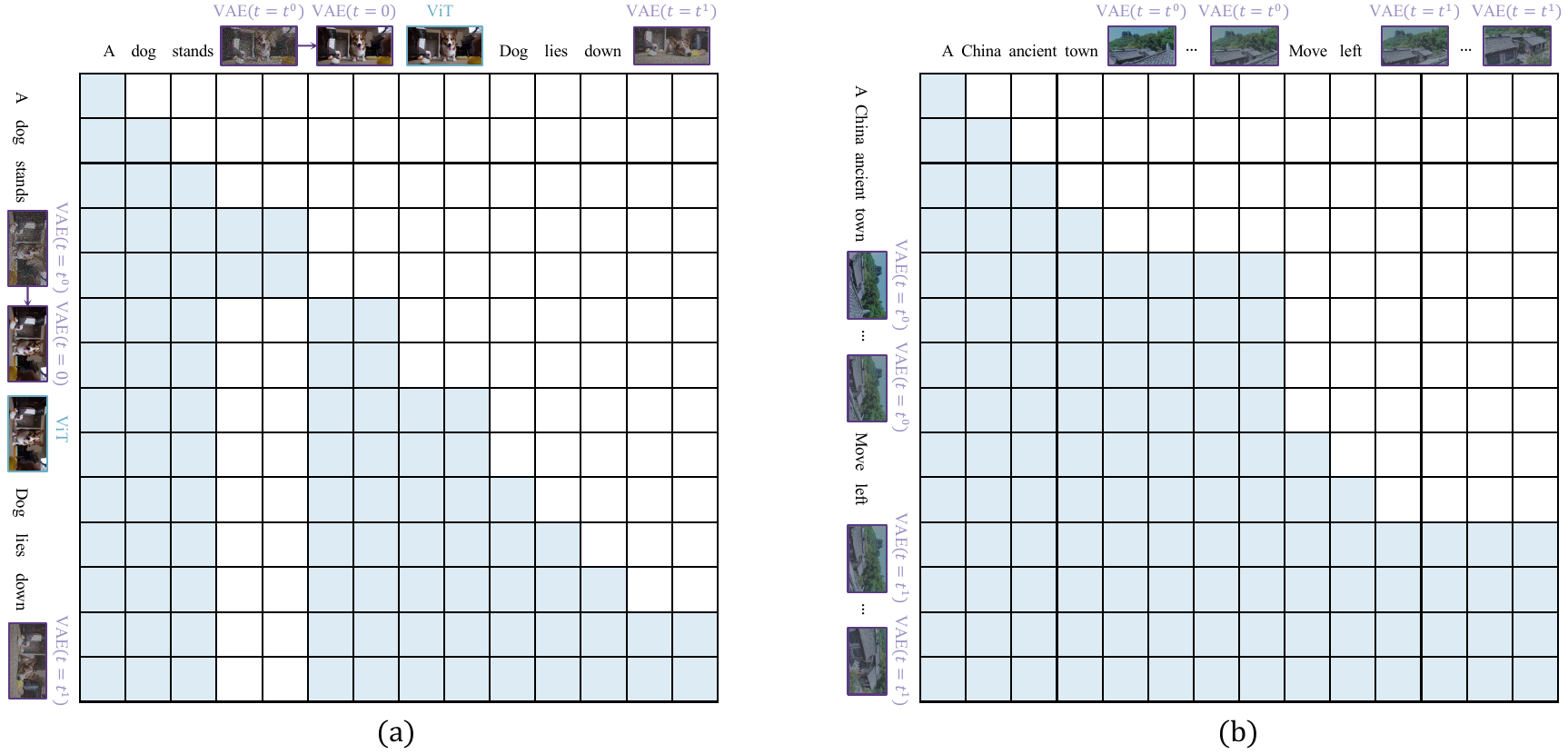}
\caption{\textbf{Causal mask in \ourmodel{} during training.} \textit{VAE} and  \textit{ViT} denote VAE features and ViT features, respectively.  \textit{t} is the noise timestep and t=0 means no noise. For each individual image, we apply full attention within its own VAE and ViT features. (a) During interleaved image-text generation, each image attends exclusively to the clean (noise-free) VAE and ViT tokens of preceding images (if present). (b) For interleaved multi-image or video clip generation, we adopt the diffusion forcing strategy~\cite{chen2024diffusion}, conditioning each image on noisy representations of preceding images. Additionally, to enhance generation consistency, we randomly group consecutive images and apply full attention within each group.  }
\label{fig:attention}
\end{figure}

We implement the generalized causal attention with {PyTorch} FlexAttention~\cite{flexattention}, achieving a \mbox{$\sim\!\!2\times$} speed‑up over naive scaled‑dot‑product attention.  
During inference, the generalized causal structure allows us to cache key-value (KV) pairs of the generated multimodal context and thus accelerate multimodal decoding.  
Only the KV pairs of {clean} VAE tokens and ViT tokens are stored; once an image is fully generated, the corresponding {noised} VAE tokens in the context are replaced by their clean counterparts.  
To enable classifier‑free guidance~\cite{ho2021classifier} in interleaved inference, we randomly drop text, ViT, and clean VAE tokens with probabilities 0.1, 0.5, and 0.1, respectively.
An illustration of the generalized casual attention is shown in \Cref{fig:attention}.

\subsection{Transformer Design}

\begin{figure*}[!t]
\begin{subfigure}[b]{0.95\textwidth}
    \centering
    \includegraphics[width=\textwidth]{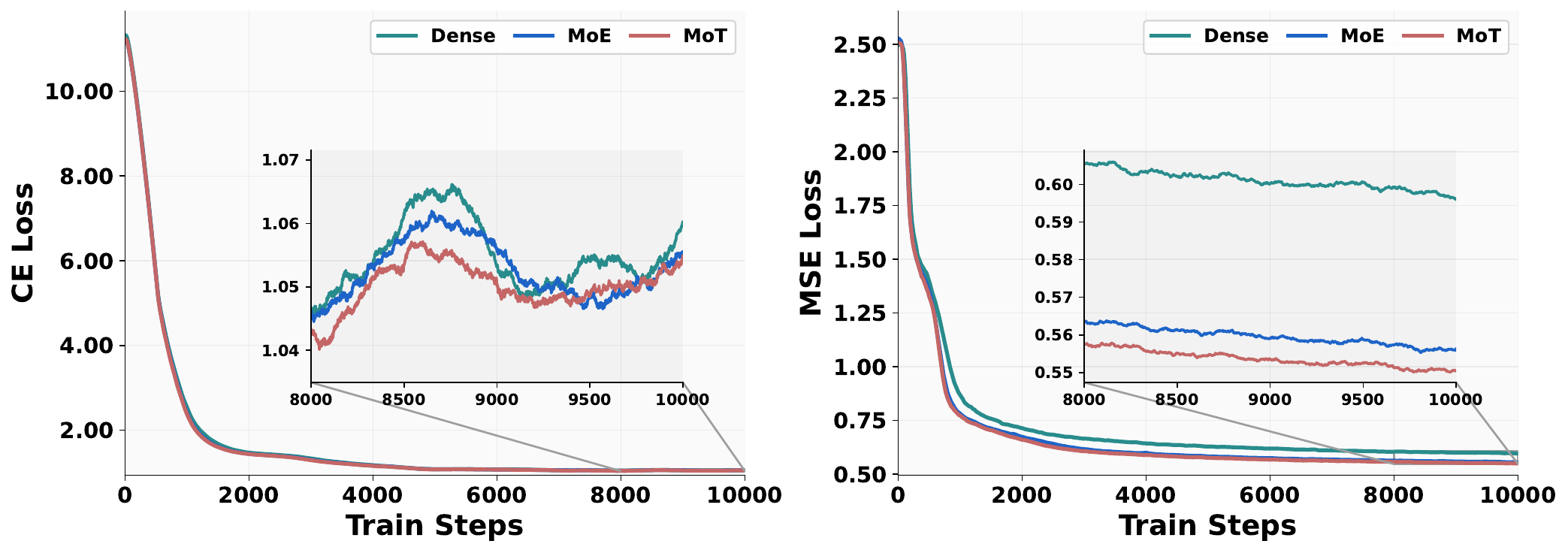}
\end{subfigure}

\caption{\textbf{Loss curves of various designs.} CE loss and MSE loss are computed on multimodal understanding and generation tasks, respectively. Ablation experiments are carried out on a 1.5B LLM. The sampling ratio for generation and understanding data is set at 4:1.}
\label{fig:arch_loss}
\end{figure*}

Following the principle of the {Integrated Transformer} solution, we compare several transformer variants: the standard {Dense Transformer}, a {Mixture‑of‑Experts} (MoE) transformer, and a {Mixture‑of‑Transformers} (MoT) architecture.
\begin{itemize}
    \item \textbf{MoE variant}: we duplicate only the feed‑forward network (FFN) in each Qwen2.5 LLM block as the initialization of the {generation expert}.
    \item \textbf{MoT variant}: we duplicate all trainable parameters of Qwen2.5 LLM to create a full-size generation expert. This type of architecture has been adopted by existing works \cite{shi2024llamafusion,mot}.
\end{itemize}
Both MoE and MoT in our model use hard routing: the newly replicated generation expert exclusively processes VAE tokens, while the original parameters—the understanding expert—handle text and ViT tokens, following the strategy of the Qwen‑VL series~\cite{qwen2vl,qwen2.5-vl}.
Although the MoE and MoT architectures increase the total parameter count by approximately twofold compared to the dense baseline, all three model variants have identical FLOPs during both training and inference.

We conduct a controlled experiment on 1.5B Qwen‑2.5 LLM, maintaining identical hyper-parameters and data configurations to isolate the transformer architecture as the sole variable. As illustrated in \Cref{fig:arch_loss}, the MoT variant consistently outperforms both the dense and MoE designs, with the gap being most pronounced on the multimodal generation task.
The MSE loss (generation) exhibits a smooth, monotonically decreasing trajectory, where MoT not only converges fastest but also attains the lowest final loss. In contrast, the CE loss (understanding) exhibits greater step‑to‑step fluctuations—an expected consequence of interleaving heterogeneous data—yet MoT still maintains the best performance in general. 
These findings highlight the clear advantage of decoupling the parameters devoted to generation from those optimized for understanding, which suggests the two objectives may steer the model toward distinct regions of the parameter space—at least at the 1.5B scale examined here. In short, allocating separate capacity for multimodal understanding and generation can mitigate optimization challenges arising from competing modality-specific learning objectives.

\section{Data}
As data define the knowledge boundaries of large foundational models, BAGEL is trained on a diverse set of datasets spanning multiple modalities—including language, image, video, and web data—enabling it to perform multimodal reasoning, in-context prediction, physical dynamics modeling, and future frame prediction, all through a unified multimodal interface. In addition to standard vision-language (VLM), text-to-image (T2I), and large-scale language modeling (LLM) datasets, we build new vision-text interleaved datasets from web and video sources to further enhance the model’s ability for sequential multimodal reasoning. 
In \Cref{tab:training_data_details}, we summarize the scale and composition of our training data across different modalities.
In the following sections, we detail our dataset sources, preparation protocols, and data mixing strategies.

\begin{table}[ht]
\small
\centering
\caption{\textbf{Data statistics for \ourmodel{}.} Since data are randomly sampled during pre-training, the dataset size does not directly correspond to the total number of seen tokens.  Multimodal interleaved data is highlight in \colorbox{lightergray}{gray}. \vspace{-15pt}}
\label{tab:training_data_details}
\begin{tabular}{l|cc}
\toprule
\textbf{Data Source} & \textbf{\# Data (M) }  & \textbf{\# Tokens (T)} \\
\hline
Text Data     &  400  &   0.4 \\
Image-Text-Pair Understanding Data  & 500 & 0.5 \\
Image-Text-Pair Generation Data  &  1600  & 2.6 \\
\rowcolor{lightergray}
Interleaved Understanding Data  &  100  & 0.5 \\ 
\rowcolor{lightergray}
Interleaved Generation Data: Video  & 45   & 0.7 \\ 
\rowcolor{lightergray}
Interleaved Generation Data: Web  & 20    & 0.4 \\ 
\bottomrule
\end{tabular}
\end{table}

\subsection{Text Only Data}
To maintain the language modeling capabilities of the underlying LLM, we supplement our training corpus with a collection of high-quality text-only data. The data are curated to support broad linguistic coverage and enable strong reasoning and generation abilities across general-purpose text tasks.

\subsection{Vision-Text Paired Data}

Text-image paired data plays a central role in multimodal learning, providing large-scale visual supervision for both vision-language models (VLMs)~\cite{li2025llavaonevision, qwen2vl} and text-to-image (T2I) generation~\cite{flux, imagen, sdxl, dalle3}. 
In our setup, we organize vision-text paired data into two subsets based on their downstream usage: one for VLM pre-training and one for T2I generation.

\textbf{VLM Image-Text Pairs.} We utilize large-scale image-text pairs for VLM training, covering a broad range of visual concepts and primarily sourced from web alt-text and captions. The data have undergone CLIP-based similarity filtering, resolution and aspect ratio constraints, text length checks, and deduplication to ensure quality and diversity. 
To address long-tail distributions, concept-aware sampling is applied to improve coverage of rare categories. In addition, structured supervision from OCR documents, charts, and grounding annotations is included to enhance the model’s capabilities in reading and spatial understanding.

\textbf{T2I Image-Text Pairs.} We incorporate high-quality image-text pairs, as well as minimal synthetic data from existing T2I models~\cite{flux,SD3}. These data feature not only diverse caption styles such as artistic, textual, and surreal captions, but also high-quality images that are filtered for clarity, structural integrity, and semantic diversity. Together, these examples enhance the visual quality and stylistic variety of our T2I training corpus.

\subsection{Vision-Text Interleaved Data}
While vision-text paired data provides useful supervision, it falls short in supporting complex in-context reasoning involving multiple images and intermediate text. Models trained on such data often struggle to capture visual and semantic relationships across modalities, resulting in less coherent generations.
To address these limitations, we incorporate large-scale vision-text interleaved data into training. For improving multimodal understanding, we utilize VLM interleaved datasets. For visual generation, we introduce a unified protocol for constructing vision-text interleaved data by combining diverse sources to support richer multimodal interactions, as detailed below.

\subsubsection{Data Source}

To comprehensively cover diverse real-world scenarios with scalable data supply, our training corpus integrates two primary sources that provide sufficient knowledge for multimodal reasoning: \textit{video data} and \textit{web data}.

\textbf{Video data} offers rich world knowledge by capturing temporal and spatial dynamics directly from the real world—the largest and most natural simulator. It preserves fine-grained visual details, maintains identity consistency across frames, and models complex motion, making it particularly effective for tasks such as image editing, navigation, and 3D manipulation.
We construct our video dataset using publicly available online video resources, as well as two open-source datasets: Koala36M~\cite{wang2024koala}, which provides large-scale instructional and interaction-rich content, and MVImgNet2.0~\cite{han2024mvimgnet2}, which contains objects captured from varying camera viewpoints to support multi-view spatial understanding.

\textbf{Web data} captures complex real-world multimodal structures and offers diverse knowledge spanning a wide range of domains. It includes naturally interleaved resources such as illustrated encyclopedic articles, step-by-step visual tutorials, and other richly grounded documents. 
This interleaved format offers rich supervision for training models to perform multimodal reasoning.
We build upon OmniCorpus~\cite{li2024omnicorpus}, a large-scale dataset preprocessed from Common Crawl~\cite{commoncrawl}, which provides a vast collection of web documents with interleaved text and images. 
We additionally include open-source image editing datasets as structured interleaved data~\cite{wei2024omniedit, zhao2024ultraedit, ge2024seedxedit, hui2024hqedit, xiao2024omnigen, bai2024humanedit}, which teach fine-grained editing behaviors and enhance the model’s ability for precise multimodal reasoning and step-by-step generation.

\subsubsection{Data Filter}

\begin{table}[t]
\small
\centering
\caption{\textbf{Quality filtering rules are applied to web documents, with each filter type accompanied by its specific filtering threshold or method.} \vspace{-15pt}}
\label{tab:web_filtering}
\begin{tabular}{l|p{12cm}} 
\toprule
\textbf{Filter Type} & \textbf{Description} \\
\midrule
UI removal & Remove images whose URLs contain substrings such as \texttt{icon} or \texttt{widget} \\
Resolution & Require width and height within [150, 20000], and aspect ratio within [1/2, 2] \\
Image clarity & Remove blurry or low-quality images using a clarity operator \\
Text density & Discard document-style images with over 100 OCR-detected text tokens \\
Relevance & Remove redundant or irrelevant images based on CLIP similarity \\
Doc. trimming & Remove unrelated headers and footers via an LLM \\
Image quantity & Keep documents with 3–8 images for balanced context \\
\bottomrule
\end{tabular}
\end{table}

\textbf{Data Filtering for Video Data.}
We follow T2V video processing pipelines~\cite{seawead2025seaweed} protocol to preprocess videos into high-quality training clips through temporal splitting, spatial cropping, and quality filtering. Videos are first segmented into short, coherent clips using lightweight shot detection, with related segments optionally merged based on visual similarity. We then remove black borders and overlays such as logos or text using crop detection and frame-level bounding box aggregation. To ensure quality, we filter clips by length, resolution, clarity, and motion stability, and deduplicate using CLIP-based similarity. This process yields a clean and diverse video dataset suitable for multimodal training.

\textbf{Data Filtering for Web Data.}
To curate high-quality interleaved data from a large corpus, we design a two-stage filtering pipeline targeting documents such as tutorials, encyclopedic entries, and design content, where text and images exhibit strong semantic alignment. Inspired by DeepSeekMath~\cite{shao2024deepseekmath}, we first apply a lightweight topic selection process: LLMs are prompted to classify a small subset of documents, and the resulting labels are used to train fastText~\cite{joulin2016fasttext} classifiers for efficient large-scale inference. The selected data are then passed through the LLM classifier again for fine-grained filtering. We adopt the 14B variant of Qwen2.5 models~\cite{qwen2.5} for its balance of performance and efficiency. 
To further improve data quality, we apply a set of rule-based filters targeting image clarity, relevance, and document structure, as summarized in \Cref{tab:web_filtering}.

\begin{figure*}[!thb]
    \centering
    \begin{minipage}[b]{1.0\textwidth}
        \centering
        \begin{subfigure}[b]{0.45\textwidth}
            \centering
            \includegraphics[height=5.0cm]{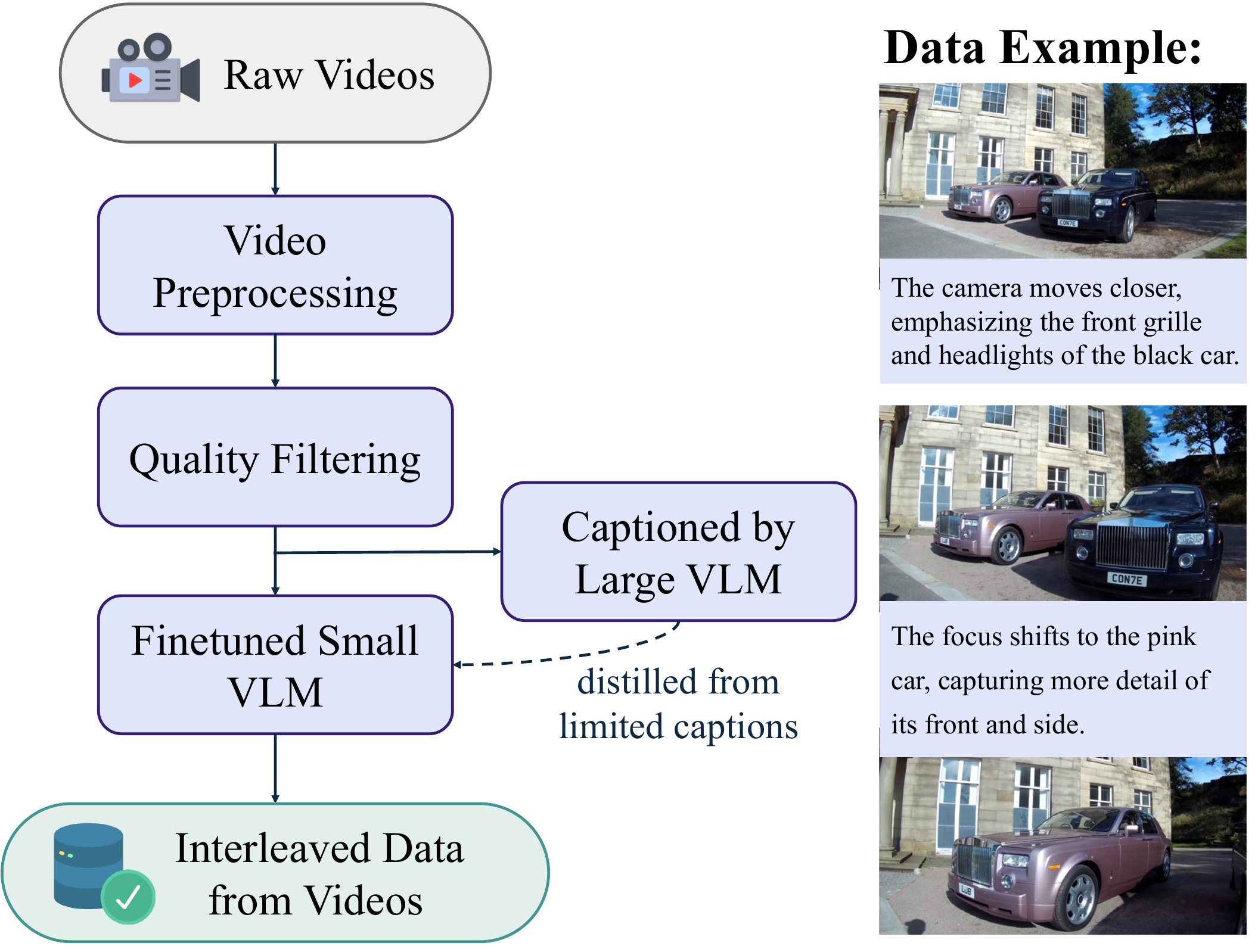}
            \caption{Data pipeline for interleaved data from videos.}
            \label{fig:video_pipeline}
        \end{subfigure}
        \hfill
        \begin{subfigure}[b]{0.53\textwidth}
            \centering
            \includegraphics[height=5.0cm]{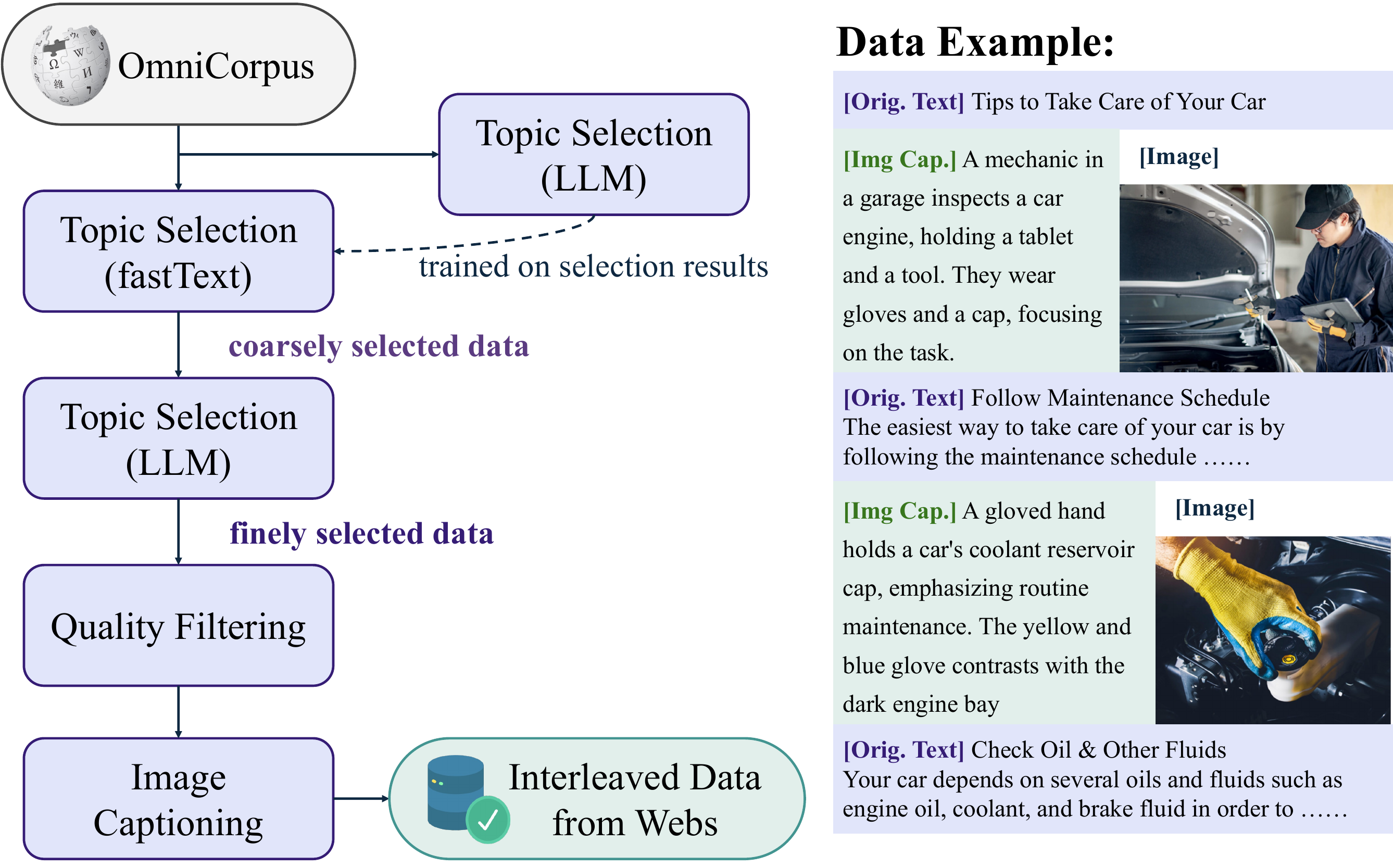}
            \caption{Data pipeline for interleaved data from webs.}
            \label{fig:web_pipeline}
        \end{subfigure}
    \end{minipage}
    \vspace{-15pt} 
   \caption{\textbf{Interleaved data construction pipelines.}
(a) We construct interleaved video data by preprocessing and filtering raw videos, then generating temporally grounded captions with a small VLM distilled from limited outputs of a large VLM.
(b) For web data, we build on OmniCorpus~\cite{li2024omnicorpus} and perform a two-stage topic selection followed by quality filtering and captioning to produce structured sequences. Data examples from both pipelines are shown.
\vspace{-15pt}}

    \label{fig:data_pipeline}
\end{figure*}

\subsubsection{Data Construction}

\textbf{Interleaved Data from Videos.}
To construct image-text interleaved sequences from video, we generate textual descriptions of visual changes between consecutive frames—capturing object motion, action transitions, and scene shifts. These inter-frame captions serve as temporal supervision for learning visual dynamics.
While large VLMs can produce high-quality change descriptions, their inference cost limits scalability. We instead distill a lightweight captioning model based on Qwen2.5-VL-7B~\cite{qwen2.5-vl}, finetuned on a small set of high-quality inter-frame examples. To reduce hallucination, we cap the caption length at 30 tokens. For each video clip, we sample an average of four frames and generate captions for each frame pair, resulting in 45 million temporally grounded interleaved sequences.
\Cref{fig:video_pipeline} illustrates the data pipeline along with an example.

\textbf{Interleaved Data from Webs.}
To construct high-quality interleaved sequences from web documents, we aim to reduce the difficulty of image generation caused by weak alignment between images, their accompanying text, and surrounding visual context. To provide more localized and relevant cues for each image, we adopt a caption-first strategy: for each image, we generate a concise description using Qwen2.5-VL-7B~\cite{qwen2.5-vl} and insert it directly before the image as a conceptual scaffold. This enables the model to form a conceptual draft of the target image-grounded in both preceding context and the inserted caption—before generating it.
By generating the caption to guide what the model should expect in the image, this approach mitigates issues caused by loosely related or ambiguous inputs. Additionally, we rewrite inter-image text segments exceeding 300 tokens using an LLM summarizer to improve contextual density. These steps yield a cleaner and more structured dataset of 20 million interleaved web documents.
Data pipeline and examples is shown in \Cref{fig:web_pipeline}.

\vspace{-2pt}
\subsubsection{Reasoning-Augmented Data}
Inspired by recent models like O1~\cite{jaech2024o1} and DeepSeek-R1~\cite{guo2025r1}, we leverage long-context Chain-of-Thoughts data for multimodal understanding. Moreover, we hypothesize that introducing a language-based reasoning step before image generation helps clarify visual goals and improve planning. To explore this, we construct 500k reasoning-augmented examples, covering four categories based on the structural relation between input and output: text-to-image generation, free-form image manipulation, and abstract edits.

\textbf{Text-to-Image generation.} We begin by manually crafting a set of brief and ambiguous T2I queries, each paired with simple generation guidance. Using in-context learning, we prompt Qwen2.5-72B~\cite{qwen2.5} to generate additional query-guidance pairs and corresponding detailed prompts, which are then passed to FLUX.1-dev~\cite{flux} to produce target images. This process yields training triplets of query, reasoning trace (guidance $+$ detailed prompt), and image, enabling models to ground image generation in language-based reasoning.

\textbf{Free-form image manipulation.}
We generate reasoning-augmented examples by prompting a VLM with the source image, target image, user query, and a reasoning trace example from DeepSeek-R1~\cite{guo2025r1}. The R1 example is generated by conditioning on the source and target captions, user query, and a reasoning instruction. The VLM prompt for the reasoning trace generation is demonstrated in \Cref{tab:prompt_edit} and \Cref{tab:prompt_wmedit}.
We sample source and target image pairs primarily from two sources: open-source editing datasets such as OmniEdit~\cite{wei2024omniedit}, and interleaved video data, which provide a rich set of naturally occurring edit scenarios characterized by substantial motion, viewpoint variations, and human interactions while preserving spatial-temporal coherence.

\textbf{Conceptual Edits.} 
Conceptual edits target cases where image manipulation requires high-level conceptual reasoning rather than simple local pixel modifications, such as transforming an object into a design sketch. For these tasks, we use the web interleaved dataset, sampling candidate image pairs from each sequence and applying a three-stage VLM pipeline to construct high-quality QA examples.
First, given a sequence of images, we prompt the VLM to identify a plausible input-output pair. Next, we prompt the model to generate a corresponding textual question based on the selected pair. Finally, we use the VLM to assess the quality of the question and its alignment with the input and output images, filtering out low-quality examples. Accepted examples are then passed to the VLM, prompted with a reasoning trace example from DeepSeek-R1~\cite{guo2025r1}, to produce grounded explanations of the intended transformation, as shown in \Cref{tab:prompt_conceptedit}. This setup helps the model learn to interpret complex visual goals from diverse textual instructions.

\begin{table}[b!]
\small
\centering

\caption{\textbf{Training recipe of BAGEL.} Multimodal interleaved data is highlight in \colorbox{lightergray}{gray}.}
    \label{tab:training_recipe}
\begin{tabular}{l|cccc}
\toprule
 & \textbf{Alignment} & \textbf{PT} & \textbf{CT} & \textbf{SFT} \\
\hline
\textbf{Hyperparameters} \\
Learning rate  & $1\times10^{-3}$ & $1.0\times10^{-4}$ & $1.0\times10^{-4}$ & $2.5\times10^{-5}$ \\
LR scheduler   & Cosine & Constant & Constant & Constant \\
Weight decay   & 0.0  & 0.0 & 0.0 & 0.0 \\
Gradient norm clip  & 1.0 & 1.0 & 1.0 & 1.0 \\
Optimizer       & \multicolumn{4}{c}{AdamW ($\beta_1=0.9$, $\beta_2=0.95$, $\epsilon=1.0 \times 10^{-15}$)} \\
Loss weight (CE : MSE)  & - & 0.25 : 1 & 0.25 : 1 & 0.25 : 1 \\
Warm-up steps   & 250 & 2500 & 2500 & 500 \\
Training steps  & 5K & 200K & 100k & 15K \\
EMA ratio            & - & 0.9999 & 0.9999 & 0.995 \\
Sequence length per rank (min, max) & (32K, 36K) & (32K, 36K) & (40K, 45K) & (40K, 45K) \\
\# Training seen tokens & 4.9B & 2.5T & 2.6T & 72.7B \\
Max context window  & 16K & 16k & 40k & 40k \\
Gen resolution (min short side, max long side)      & - & (256, 512) & (512, 1024) & (512, 1024) \\
Und resolution (min short side, max long side)      & (378, 378) & (224, 980) & (378, 980) & (378, 980) \\
Diffusion timestep shift & - & 1.0 & 4.0 & 4.0 \\
\midrule
\textbf{Data sampling ratio} \\
Text                            & 0.0 & 0.05 & 0.05 & 0.05 \\
Image-Text pair (T2I)           & 0.0 & 0.6 & 0.4 & 0.3 \\
Image-Text pair (I2T)           & 1.0 & 0.1 & 0.1 & 0.05 \\
\rowcolor{lightergray}
Interleaved understanding       & 0.0 & 0.1 & 0.15 & 0.2 \\
\rowcolor{lightergray}
Interleaved generation: video   & 0.0 & 0.1 & 0.15 & 0.2 \\
\rowcolor{lightergray}
Interleaved generation: web     & 0.0 & 0.05 & 0.15 & 0.2 \\
\bottomrule
\end{tabular}
\end{table}

\section{Training}

As shown in \Cref{tab:training_recipe}, we adopt a multi-stage training strategy using a dynamic mixture of the curated data described above—specifically, an Alignment stage for initializing the VLM connector, a Pre-training stage for large-scale pre-training, a Continued Training stage for increased resolution and interleaved data ratio, and a Supervised Fine‑tuning stage for high-quality fine-tuning:

\begin{itemize}
    \item \textbf{Stage: Alignment}. In this stage, we align the SigLIP2 ViT encoder with the Qwen2.5 LLM by training only the MLP connector while keeping the vision encoder and the language model frozen. Only image–text pair data are used during this stage to perform image captioning, where each image is resized to a fixed resolution of $378 \times 378$ to match the input size of the pre-trained SigLIP2.
    
    \item \textbf{Stage: Pre‑training (PT).} During this stage, we add QK-Norm to the LLM and all model parameters except those of the VAE are trainable.  
    The training corpus comprises 2.5T tokens, consisting of text, image–text pairs, multimodal conversation, web‑interleaved, and video‑interleaved data.  
    We adopt a native‑resolution strategy for both multimodal understanding and generation, with restrictions on the maximum long side and minimum short side of each image.
    
    \item \textbf{Stage: Continued Training (CT).} Compared with PT, we increase the visual input resolution in the CT stage, which is important for both multimodal generation and understanding performance.  
    We further strategically increase the sampling ratio of interleaved data to emphasize learning cross-modal reasoning, as the model’s core understanding and generation capabilities become more stable and reliable.  
    The CT stage consumes approximately 2.6T tokens.
    
    \item \textbf{Stage: Supervised Fine‑tuning (SFT).} In the SFT stage, for multimodal generation we construct a high‑quality subset from the image–text‑pair dataset and the interleaved‑generation dataset. For multimodal understanding, we filter a subset from the LLaVA‑OV~\cite{li2025llavaonevision} and Mammoth‑VL~\cite{guo2024mammoth} instruction‑tuning data.  
    The total number of training tokens at this stage is 72.7billion. 
    
\end{itemize}
\vspace{5pt}

For all training stages, we use the AdamW~\cite{loshchilov2017decoupled} optimizer with $\beta_{1}=0.9$, $\beta_{2}=0.95$.  
Inspired by~\cite{molybog2023theory}, we set $\epsilon = 1.0\times10^{-15}$ to suppress loss spikes.  
When increasing the resolution for generation, we also increase the diffusion timestep from 1.0 to 4.0 to ensure a proper noise‑level distribution.  
We adopt a constant learning rate for the PT, CT, and SFT stages so that we can easily scale the training data without restarting the training process~\cite{hu2024minicpm}. 
To ensure load balance among different ranks, we pack the sequences on each rank into a narrow length range (32K to 36K tokens for Alignment and PT, 40K to 45K tokens for CT and SFT).

Unlike the pre‑training of standalone VLMs or T2I models, unified multimodal pre‑training requires careful tuning of two key hyper‑parameters—the data‑sampling ratio and the learning rate—to balance signals from understanding and generation tasks.  
Below, we describe the empirical insights that guided these choices, which in turn shaped the training protocol summarized in \Cref{tab:training_recipe}.

\begin{figure*}[!t]
\begin{subfigure}[b]{0.93\textwidth}
    \centering
    \includegraphics[width=\textwidth]{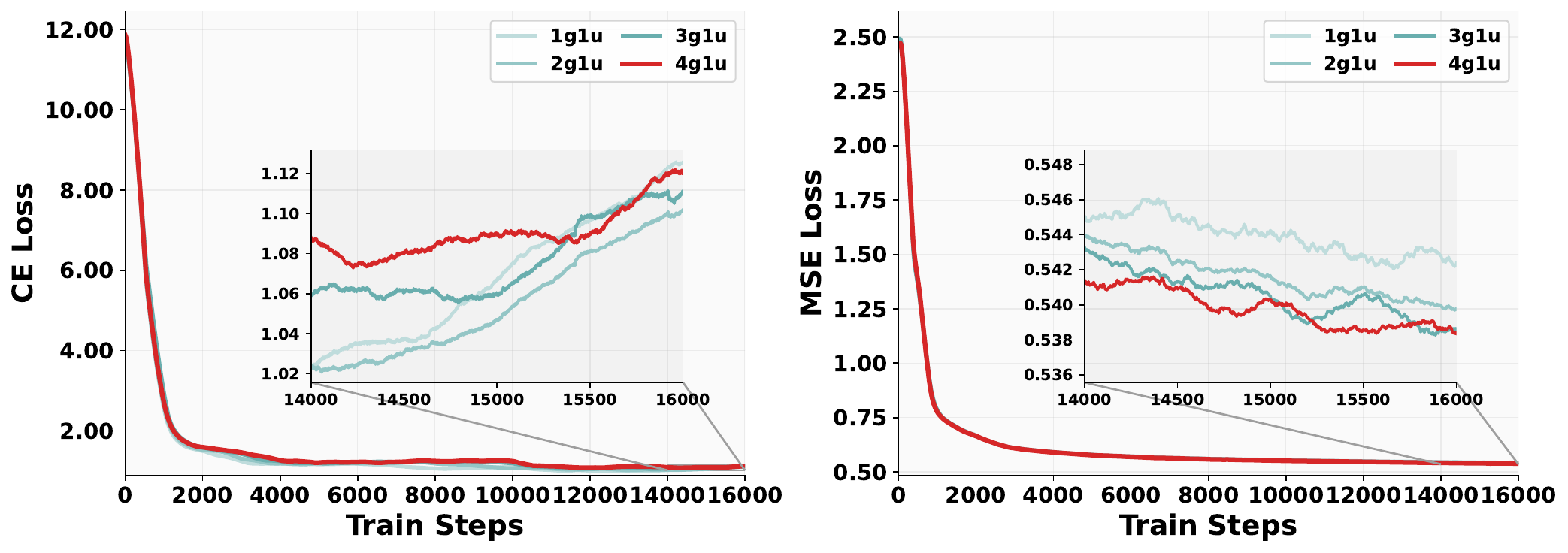}
\end{subfigure}

\caption{ \textbf{Loss curves of different data ratios.} Ablation experiments are carried out on a 1.5B LLM. "1g1u" means that the sampling ratio for generation and understanding data is set at 1:1. }
\vspace{-11pt}
\label{fig:ratio_loss}
\end{figure*}

\begin{figure*}[!t]
\begin{subfigure}[b]{0.93\textwidth}
    \centering
    \includegraphics[width=\textwidth]{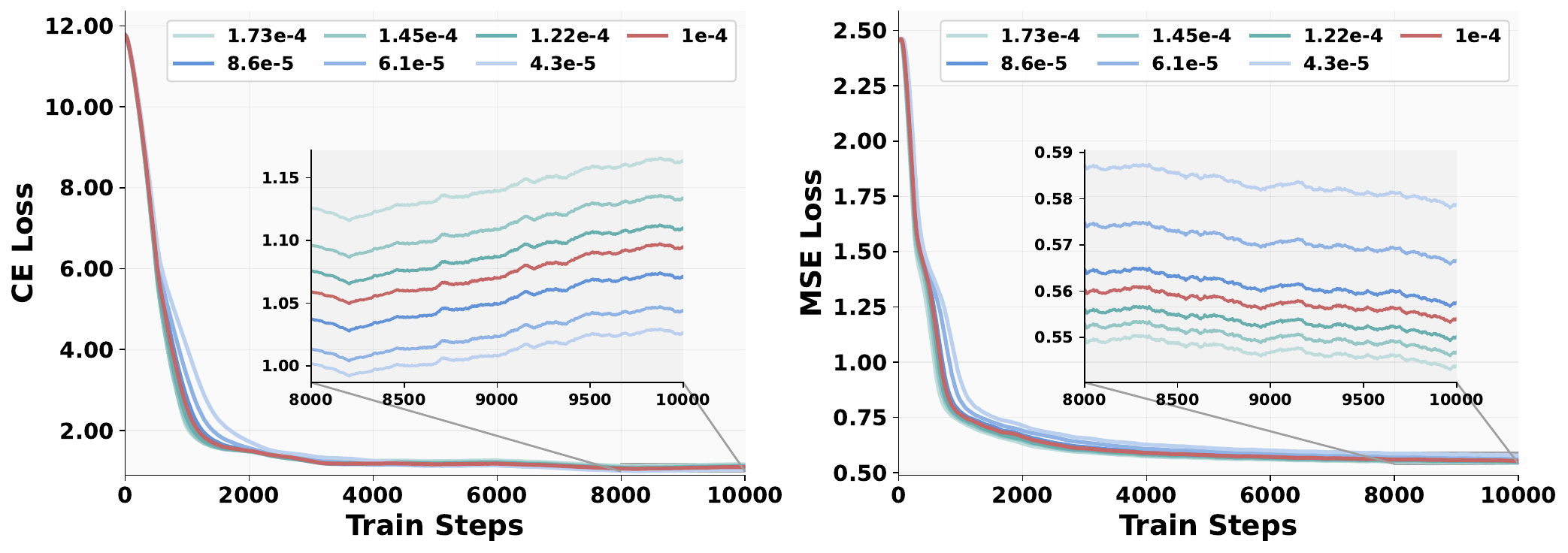}
\end{subfigure}

\caption{ \textbf{Loss curves of different learning rates.} Ablation experiments are carried out on a 1.5B LLM. The sampling ratio for generation and understanding data is set at 1:1. }
\vspace{-11pt}
\label{fig:lr_loss}
\end{figure*}

\subsection{Data Sampling Ratio}\label{sec:4.1}

To choose the sampling ratios for each data source during unified pre-training, we conducted a series of controlled studies on the 1.5B Qwen2.5 LLM~\cite{qwen2.5} by adjusting the proportion of multimodal generation data versus multimodal understanding data.
As shown in \Cref{fig:ratio_loss}, increasing the sampling ration of generation data from 50\% ("1g1u") to 80\% ("4g1u") steadily reduces the MSE loss, results in a 0.4\% absolute reduction—a considerable margin for rectified‑flow models in practice.
In contrast, the cross‑entropy (CE) loss exhibits no consistent pattern across sampling ratios; the largest observed gap, 0.07 at step 14,000 between "4g1u" and "2g1u", has negligible impact on downstream benchmarks. 
These findings suggest that generation examples should be sampled substantially more often than understanding examples—a heuristic we adopt throughout the training protocol summarized in \Cref{tab:training_recipe}.

\subsection{Learning Rate}

We next carried out a controlled experiment identical to the setup in \Cref{sec:4.1} except for the learning‑rate setup.  
As shown in \Cref{fig:lr_loss}, the two losses behave oppositely: a larger learning rate makes the MSE loss converge faster, whereas a smaller learning rate benefits the CE loss.  
To reconcile this trade‑off, we assign separate weighting factors to the two objectives, as listed in \Cref{tab:training_recipe}.

\vspace{-5pt}
\section{Evaluation}
\vspace{-5pt}
To comprehensively evaluate a unified model, we draw on established benchmarks that target well‑defined skills such as multimodal understanding, T2I generation, and classical image editing. Yet for capabilities that demand strong multimodal reasoning and complex task composition, effective evaluation strategies are still lacking. 
In the following, we first illustrate the available benchmarks used during our evaluation process,
and then introduce a new evaluation suite for free‑form image manipulation (including conceptual editing), designed to reveal a model’s proficiency in multimodal reasoning and complex compositional tasks.

\textbf{Multimodal understanding.}
We adopt six widely used benchmarks—MME \cite{fu2023mme}, MMBench ($1.0$-EN) \cite{liu2024mmbench}, MM-Vet \cite{yu2024mm}, MMMU \cite{yue2024mmmu}, MathVista \cite{lu23mathvista}, and MMVP \cite{tong2024eyes}. Collectively they offer a concise but comprehensive testbed that spans perception, cognition, and multimodal reasoning, while retaining strong discriminative power for ranking state-of-the-art models.

\textbf{Text-to-Image generation.}
We follow \cite{januspro2025,pan2025transfer} and report results on the popular GenEval~\cite{ghosh2023geneval} benchmark. We also adopt the recently proposed WISE benchmark~\cite{niu2025wise}, which offers a comprehensive assessment of complex semantic understanding and world-knowledge integration in text-to-image generation. In addition, we include qualitative comparisons with state-of-the-art models as a complement to these automatic evaluation metrics.

\textbf{Image Editing.}
We adopt GEdit-Bench \cite{liu2025step1x} as our primary evaluation suite owing to its real-world relevance and diverse set of editing tasks. Built from authentic user requests scraped from the web, GEdit-Bench closely mirrors practical editing needs. Performance is scored automatically with GPT-4.1~\cite{gpt4.1}, and we also supplement these scores with qualitative examples to provide a more nuanced assessment.

\textbf{Intelligent Image Editing.}
We propose \textit{IntelligentBench} as a proxy task for the evaluation of free-form image manipulation ability, which requires complex multimodal reasoning and task composition.
The initial release of IntelligentBench comprises 350 examples, each consisting of a question image, question text, and a reference answer image. Evaluation is performed using GPT-4o (version: gpt-4o-2024-11-20), which reviews a complete quadruplet—the question image, question text, reference answer image, and the model-generated image. The evaluation criteria include request fulfillment, visual consistency, and knowledge-grounded creativity, reflecting the benchmark’s focus on both task correctness and the depth of reasoning.
Each answer is scored on a scale from 0 to 2. The final score of a model is calculated by summing all individual scores and normalizing the total to a 100-point scale. The detailed evaluation prompt is provided in Appendix \Cref{tab:prompt_intelligentEditEval}. With the help of IntelligentBench, we can evaluate how well the model performs reasoning and integrates world knowledge for image editing.
Some showcases and qualitative results on IntelligentBench can be found in \Cref{fig:Intelligent_all}.

\vspace{-5pt}
\section{Emerging Properties}
\vspace{-5pt}

\begin{figure*}[!t]
    \centering
    \begin{minipage}[b]{1.0\textwidth}
        \centering
        \begin{subfigure}[b]{0.48\textwidth}
        
            \centering

            \includegraphics[height=4.5cm]{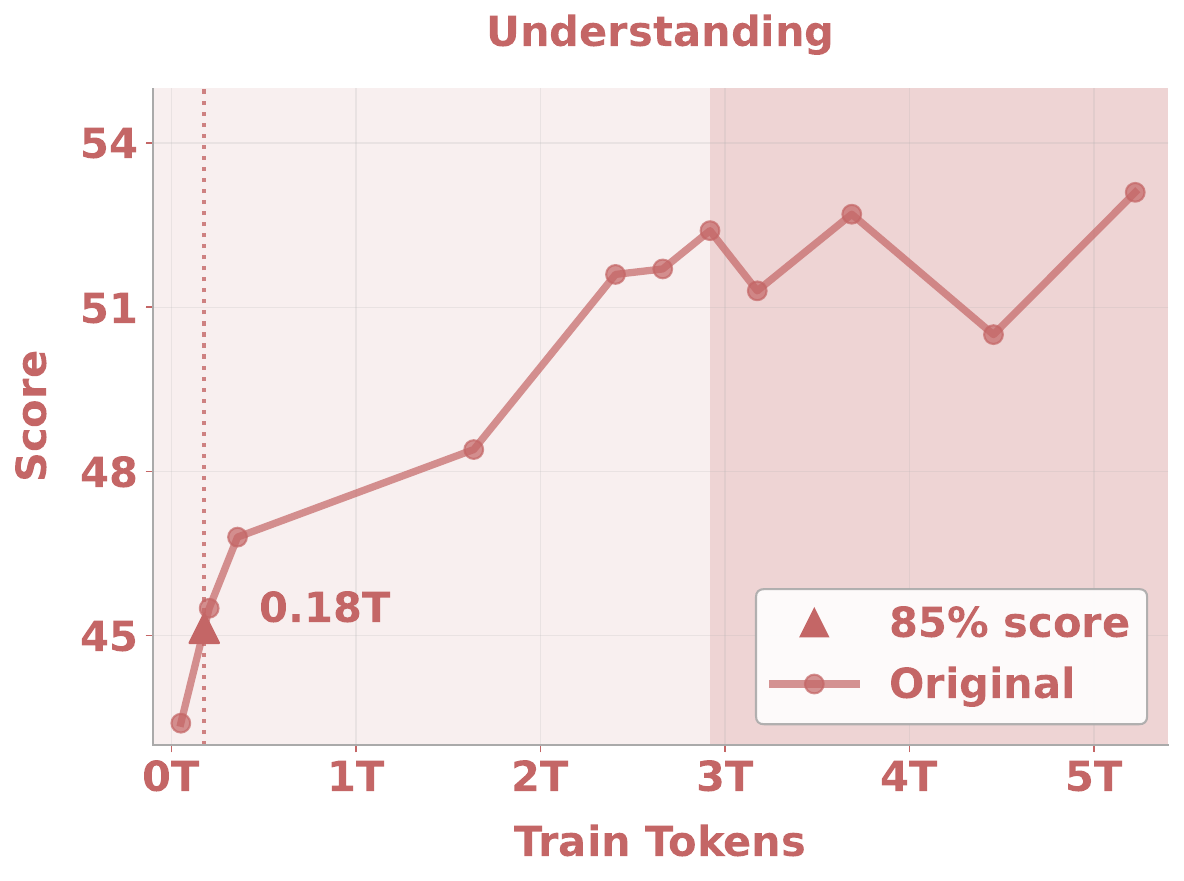}
            \caption{Average score on Image Understanding tasks.}
            \label{fig:fig1}
        \end{subfigure}
        \hfill
        \begin{subfigure}[b]{0.48\textwidth}
            \centering

            \includegraphics[height=4.5cm]{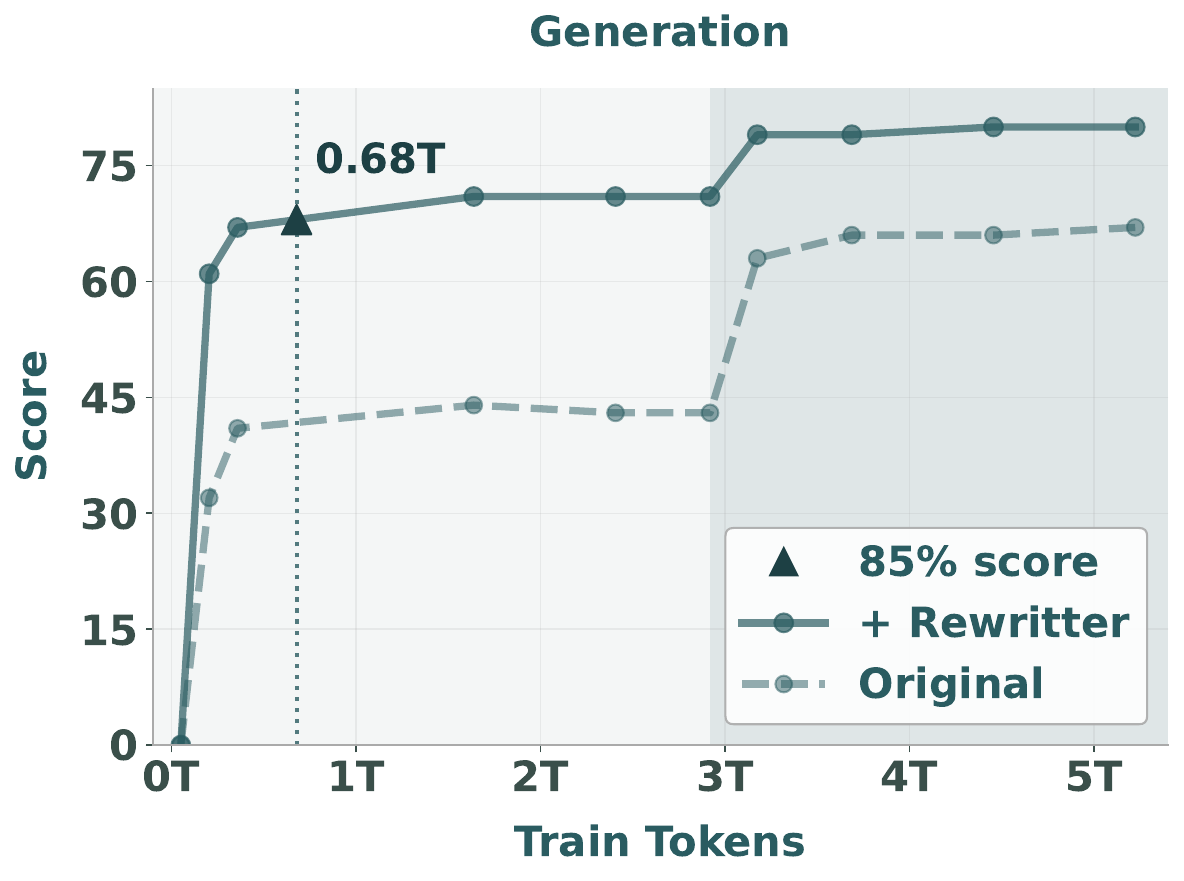}
            \caption{GenEval score on Image Generation task.}
            \label{fig:fig2}
        \end{subfigure}

        \vspace{0.4em}
        \begin{subfigure}[b]{0.48\textwidth}
            \centering

            \includegraphics[height=4.5cm]{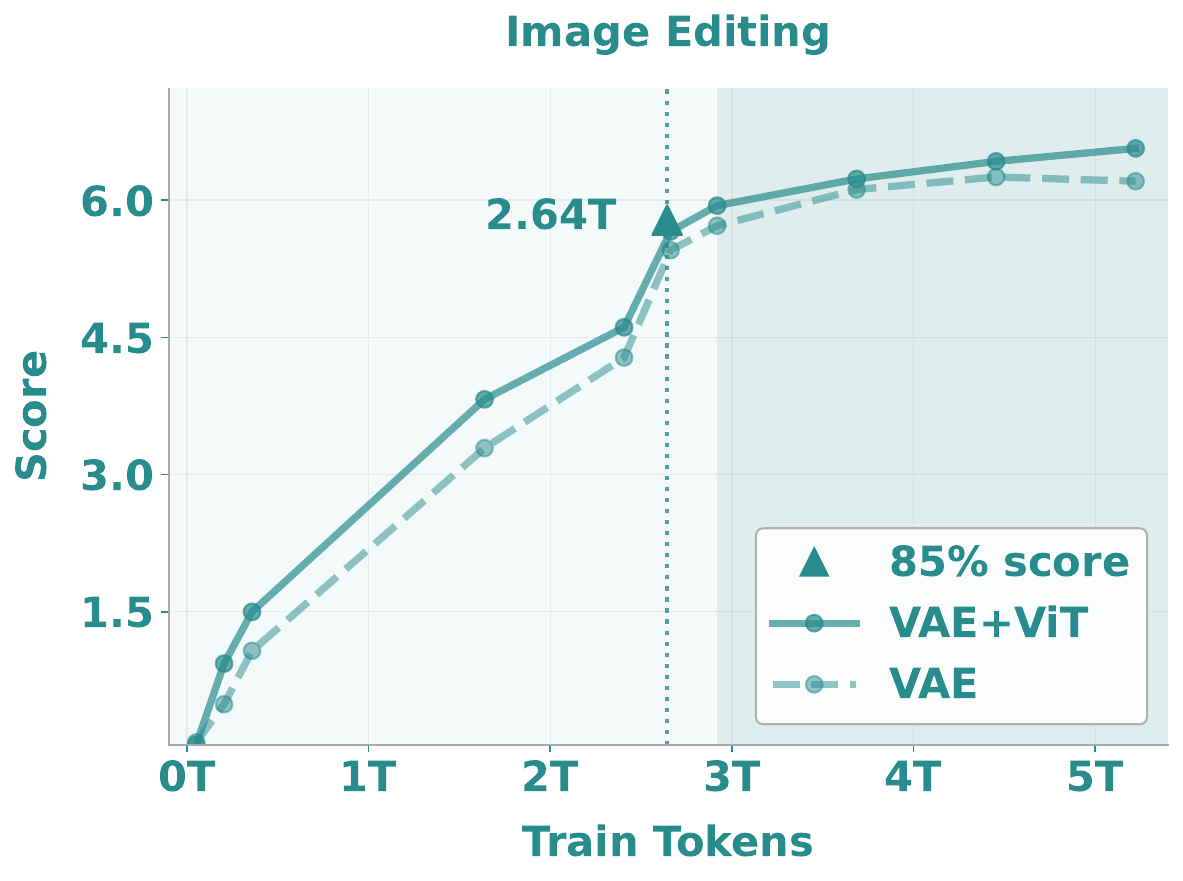}
            \caption{GEdit Overall Score on classical Image Editing task.}
            \label{fig:fig3}
        \end{subfigure}
        \hfill
        \begin{subfigure}[b]{0.48\textwidth}
            \centering

            \includegraphics[height=4.5cm]
    {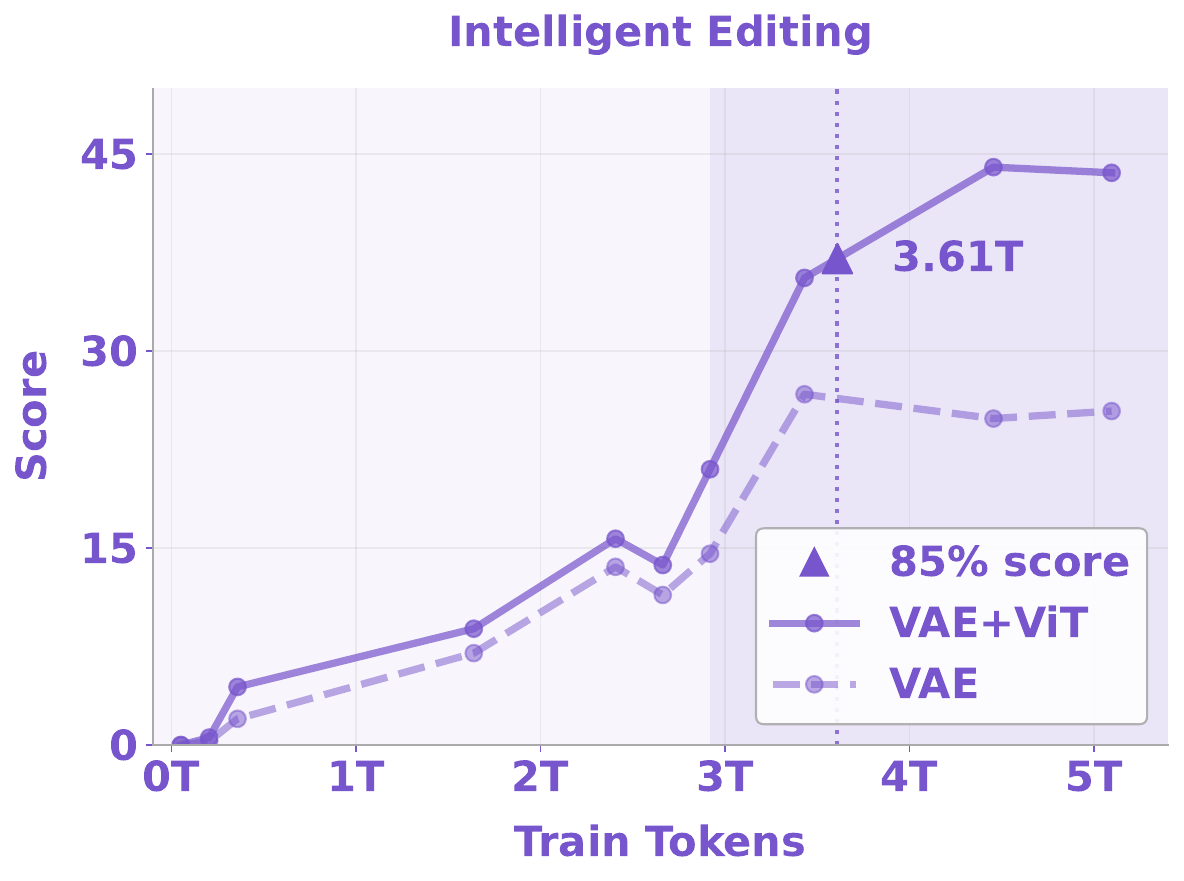}
            \caption{IntelligentBench Score on Intelligent Editing task.}
            \label{fig:fig4}
        \end{subfigure}
    \end{minipage}
    \vspace{7pt}
   \caption{\textbf{Emerging curves.} Pre-training performance curves of BAGEL on different tasks.
The lighter region represents the low-resolution pre-training stage, while the darker region indicates the high-resolution CT stage.
BAGEL demonstrates consistent performance improvements as the number of training tokens increases.
The relationship between performance and training scale can be summarized as follows:
\textbf{(i) BAGEL continues to improve }across various tasks with more training tokens;
\textbf{(ii) Different capabilities emerge at different stages}—understanding and generation abilities emerge first, basic editing follows, and intelligent editing emerges last, reflecting the increasing complexity of these tasks. \textbf{(iii) Adopting both VAE and ViT features surpasses using VAE features alone in the image editing tasks}, especially in Intelligent Editing, with a noticeable gap. This supports the idea that ViT provides important semantic context to aid generation.
Note: The average image understanding score is computed as the mean of the scores from MME-S, MMBench, MMMU, MMVet, MathVista and MMVP. All performance evaluations are conducted with BAGEL’s thinking mode disabled. }

    \label{fig:four_images}
\end{figure*}

\begin{figure*}[!ht]    
  \vspace*{-0.05\textwidth}
  \hspace*{-0.1\textwidth}
  \includegraphics[width=1.175\textwidth]{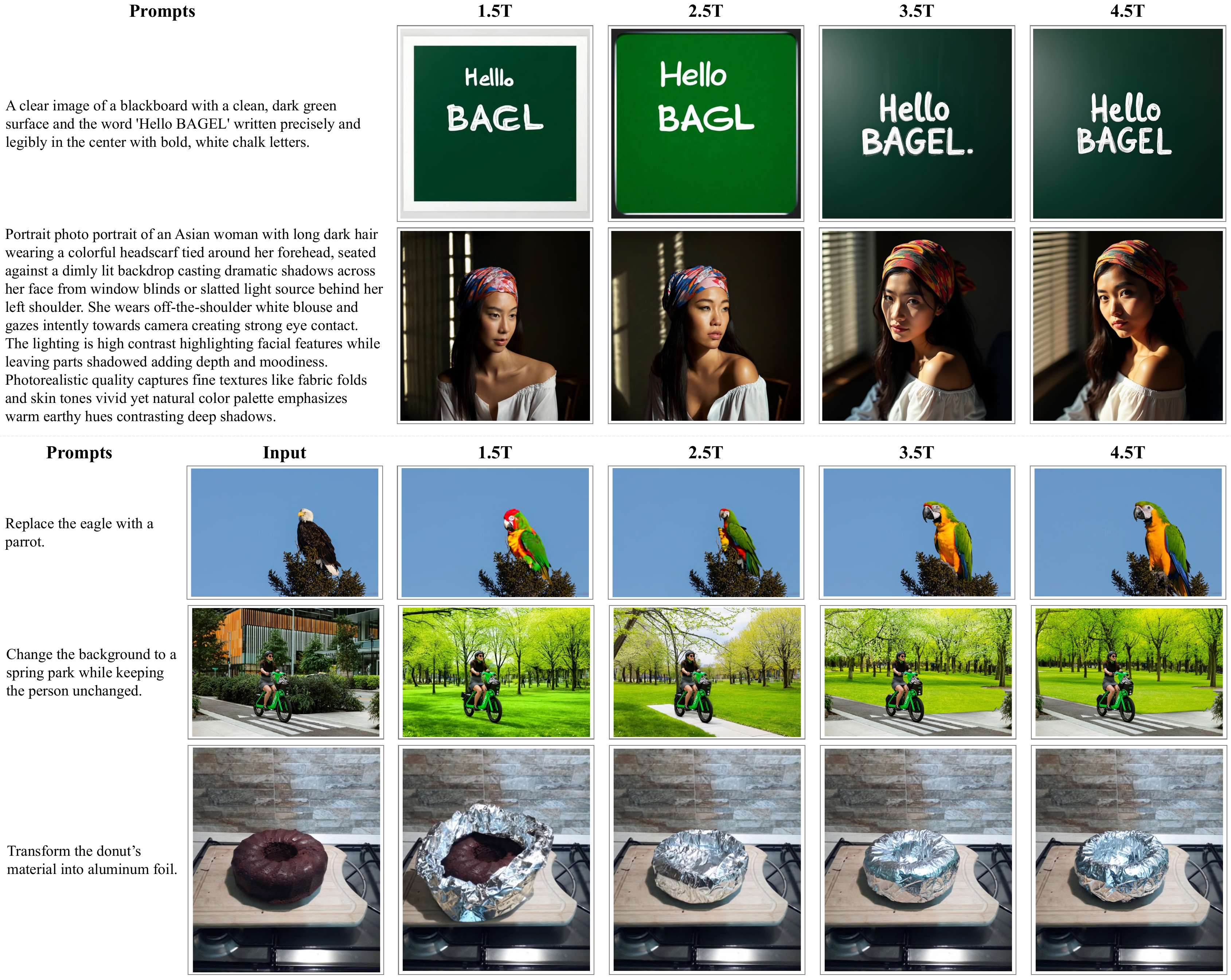}

  \caption{\textbf{Comparison of models with different amounts of training tokens.} We present cases of Text-to-Image generation and image editing.}
\label{fig:history_vis1}
\end{figure*}

\begin{figure*}[!htbp]     
  \vspace*{-0.05\textwidth}
  \hspace*{-0.1\textwidth}
  \includegraphics[width=1.175\textwidth]{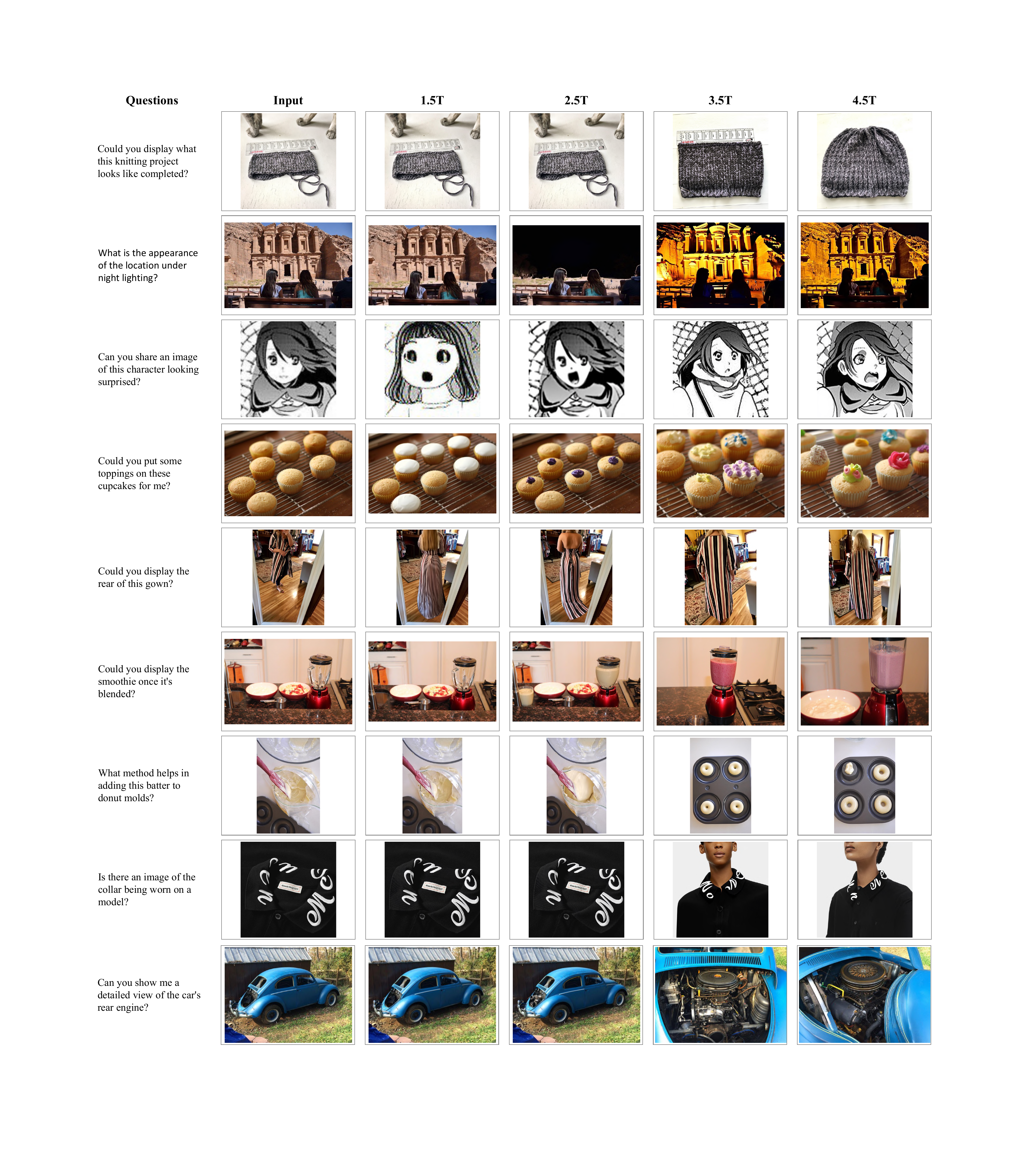}
  \caption{\textbf{Comparison of models with different amounts of training tokens.} We present cases of intelligent editing that requires strong multimodal reasoning abilities.}
\label{fig:history_vis2}
\end{figure*}

Emerging properties have been studied extensively in the context of large visual or language models~\cite{wei2022emergent,caron2021emerging}. In this work, situated within the scope of unified multimodal foundational models, 
we adopt a more focused definition for emerging properties: 

\begin{center}
\textit{An ability is emerging if it is not present in earlier training stages but is present in later pre-trainings.}
\end{center}

This qualitative shift, often referred to as a phase transition, denotes a sudden and dramatic change in model behavior that cannot be predicted by extrapolating from training loss curves~\cite{wei2022emergent}. Interestingly, we observe the similar phenomenon in unified multimodal scaling, where loss curves do not explicitly signal the emergence of new capabilities.
Therefore, we investigate the emergence of model capabilities by evaluating performance across a range of tasks on historical checkpoints. Specifically, we report the average performance on standard VLM benchmarks as a proxy for multimodal understanding, the GenEval score for generation ability, and the GEdit score and IntelligentBench score to assess the model’s capability in naive and complex multimodal reasoning, respectively.

Interestingly, different tasks demonstrate distinct learning dynamics and saturation behaviors. If we choose the number of seen tokens required to reach 85\% of peak performance as an indicator, as noted in \Cref{fig:four_images}, we find that conventional understanding and generation benchmarks saturate relatively early: at approximately 0.18T and 0.68T tokens, respectively. In contrast, editing tasks, which require both understanding and generation capabilities, exhibit slower convergence, reaching 85\% performance only after 2.64T tokens.

Most notably, the Intelligent Edit task—designed to eliminate naive edit cases and emphasize on complex multimodal reasoning—requires 3.61T tokens to reach 85\%, demonstrating a pattern akin to emergent behaviors described in~\cite{wei2022emergent}. In this setting, the model shows initially low performance that improves gradually and significantly after the 3T seen tokens. While traditional editing tasks remain largely unaffected by the resolution increase at 3T tokens, Intelligent Editing performance keeps improving significantly—from 15 to 45—tripling in later training stages and underscoring its dependence on unified multimodal reasoning. We further find that understanding ability, particularly visual input, plays a critical role in multimodal reasoning: removing the ViT tokens has minimal impact on GEdit-Bench but causes a 16\% drop in Intelligent Edit, highlighting the importance of visual-semantic reasoning in complex editing tasks.

While evaluation metrics may not linearly capture the model’s true capabilities—potentially leading to spurious signs of emergence, albeit unlikely—we further examine qualitative emerging behavior by inspecting generation outputs across different training checkpoints. As illustrated in \Cref{fig:history_vis1}, we observe trends consistent with the performance curves: generation quality is already strong before 1.5T seen tokens , with a small quality improvement after 3.0T seen tokens when trained with higher resolution. For text rendering, the ability to generate correct spell of "hello" and "BAGEL" emerge later—around 1.5T to 4.5T tokens.

The emerging behavior is also observed in the qualitative visualization of Intelligent Editing task in \Cref{fig:history_vis2}. Unlike traditional editing shown in \Cref{fig:history_vis1}, which involves only partial modifications to the input image, Intelligent Editing often requires generating entirely new concept based on multimodal reasoning. Prior to 3.5T tokens, the model tends to reproduce the input image with minimal changes—a fallback strategy when the task is not fully understood. However, after seeing 3.5T tokens, the model begins to demonstrate clear reasoning, producing coherent and semantically appropriate edits, aligning with the emergent behavior seen in \Cref{fig:four_images}.

\section{Main Results}

In this section, we present both quantitative and qualitative evaluations to examine the diverse multimodal capabilities of BAGEL.
We begin with basic abilities on established benchmarks, including image understanding in \Cref{subsec:vis_und} and image generation in \Cref{subsec:vis_gen}. We then report performance on existing image editing benchmarks and \textbf{IntelligentBench} in \Cref{subsec:editing}. 
In \Cref{subsec:thing_gen}, we explore generation and editing with explicit reasoning. In this setting, BAGEL is allowed to generate intermediate thinking steps before final outputs. We find that such reasoning significantly enhances performance. Finally, in \Cref{subsec:worldnav}, we provide qualitative visualizations that showcase BAGEL’s world modeling abilities, including world navigation and video generation.

\subsection{Image Understanding}\label{subsec:vis_und}
\begin{table*}[ht]
    \centering
    \setlength{\tabcolsep}{2pt}
    \renewcommand{\arraystretch}{1.2}
    \scriptsize
    \caption{\textbf{Comparison with state-of-the-arts on viusal understanding benchmarks.} MME-S refers to the summarization of MME-P and MME-C.
    For MoE models, we report their activate params / total params. $\dagger$: MetaQuery~\cite{pan2025transfer} adopts pre-trained model from Qwen2.5-VL~\cite{qwen2.5-vl} and freezes it during training.
    $**$: Partial results are from by MetaMorph~\cite{tong2024metamorph} or MetaQuery~\cite{pan2025transfer}.
    }
    \label{tab:sota_result_understanding}
    \begin{tabular}{clccccccccc}
        \toprule
        \textbf{Type} & \textbf{Model} & \textbf{\# LLM Params} & \textbf{MME-P$\uparrow$} & \textbf{MME-S$\uparrow$} &\textbf{MMBench$\uparrow$} & \textbf{MMMU$\uparrow$} & \textbf{MM-Vet$\uparrow$} & \textbf{MathVista$\uparrow$} & \textbf{MMVP$\uparrow$} \\
        \midrule
        \multirow{13}{*}{\rotatebox{90}{\textit{Und. Only}}}
        & InternVL2~\cite{internvl2}& 1.8B & 1440 & 1877 &  73.2  & 34.3 & 44.6 & 46.4 & 35.3 \\
        & InternVL2.5~\cite{internvl2.5}& 1.8B & - & 2138 & 74.7  & 43.6 &60.8 & 51.3 & - \\
        & Qwen2-VL\cite{qwen2vl} & 1.5B & - & 1872 & 74.9 & 41.1 & 49.5 & 43.0 & - \\
        & Qwen2.5-VL\cite{qwen2.5-vl} & 3B & - &2157 &79.1 & 53.1& 61.8& 62.3 & - \\
        & BLIP-3 \cite{xue2024xgenmmblip3familyopen} & 4B  & - & - &  76.8 & 41.1 &-& 39.6 & - \\
        & LLava-OV~\cite{li2025llavaonevision} & 7B &1580 & - &80.8 &48.8&57.5&63.2 & - \\
        & InternVL2~\cite{internvl2}& 7B & 1648 & 2210 &  81.7  & 49.3 & 54.2 & 58.3 & 51.3 \\
        & InternVL2.5~\cite{internvl2.5}& 7B & - &2344& \emph{84.6}  & 56.0 &62.8 &64.4 & - \\
        & Qwen2-VL~\cite{qwen2vl}  & 7B & - &2327 &83.0 & 54.1& 62.0& 58.2 & - \\
        & Qwen2.5-VL\cite{qwen2.5-vl} & 7B & - & \emph{2347} & 83.5 & \textbf{58.6} & \emph{67.1} & 68.2 & - \\
        & Emu$3$-Chat$^{**}$~\cite{emu3} & 8B  & 1244 & - & 58.5 & 31.6 & 37.2 & - & 36.6 \\
        & Kimi-VL~\cite{kimiteam2025kimivltechnicalreport} & 2.8B/16B & - & - & - & \emph{57.0} &66.7& \emph{68.7} & - \\
        & DeepSeek-VL2 \cite{wu2024deepseek}& 4.1B/28B  & -& - & - & 51.1 &60.0& 62.8 & - \\
        \midrule
        \multirow{15}{*}{\rotatebox{90}{\textit{Unified}}}
       & Show-o$_{512}$~\cite{show-o} & 1.3B & 1097 & - & -  & 26.7 & -&- & - \\
        & Janus~\cite{janus2024} & 1.5B& 1338 & - & 69.4& 30.5 & 34.3&- & - \\
        & Janus-Pro~\cite{januspro2025} & 1.5B& 1444 & - & 75.5 & 36.3 & 39.8 &-  & -\\
        \rowcolor{myblue}
        & \textbf{BAGEL} & 1.5B MoT & 1610 & 2183  & 79.2 & 43.2 & 48.2 & 63.4 & \emph{54.7}\\
        & ILLUME~\cite{wang2024illume} & 7B&  1445 & - &  75.1 & 38.2 &  37.0 &- & - \\
        & VILA-U$_{256}^{**}$~\cite{vila-u} & 7B & 1336 & - & 66.6 & 32.2 & 27.7 & - & 22.0 \\
        & Chameleon$^{**}$~\cite{chameleon} & 7B & -  & - & 35.7 & 28.4 & 8.3 &- & 0.0 \\
        & Janus-Pro~\cite{januspro2025} & 7B & 1567 & - & 79.2 & 41.0& 50.0&-  & -\\
        & MetaQuery-XL$^{\dagger}$~\cite{pan2025transfer} & 7B& \emph{1685} & - & 83.5 & 58.6 &66.6&- & -\\
        & LlamaFusion$^{**}$~\cite{shi2024llamafusion}  & 8B & 1604& - & 72.1& 41.7 & - &- & -\\
        & MetaMorph~\cite{tong2024metamorph} & 8B& -& - & 75.2  & 41.8 & - &- & 48.3\\
        & SEED-X~\cite{seed-x} & 13B& 1457 & - &70.1 & 35.6 &  43.0& - & -  \\
        & TokenFlow-XL~\cite{qu2024tokenflow} & 13B&  1546 & - &  68.9  & 38.7 &  40.7 &- & - \\
        & MUSE-VL~\cite{xie2024muse} & 32B & - & - & 81.8 & 50.1 & - & 55.9 & - \\ 
        \rowcolor{myblue}
        & \textbf{BAGEL} & 7B MoT & \textbf{1687} & \textbf{2388}  & \textbf{85.0} & 55.3 & \textbf{67.2} & \textbf{73.1} & \textbf{69.3}\\
        \bottomrule
    \end{tabular}
    
\end{table*}

We extensively benchmark \ourmodel{} against state-of-the-art open-source multimodal models, including both specialized visual understanding and general-purpose unified models. Our evaluation spans a diverse set of public benchmarks to ensure a comprehensive assessment of model capabilities.

The visual understanding results are summarized in \Cref{tab:sota_result_understanding}.
At a comparable activated parameter size of 7B, \ourmodel{} outperforms existing unified models in understanding tasks. For instance, it achieves significant improvements of 14.3 and 17.1 points over Janus-Pro~\cite{januspro2025} on MMMU and MM-Vet, respectively. Notably, MetaQuery-XL~\cite{pan2025transfer} relies on a frozen, pre-trained Qwen2.5-VL~\cite{qwen2.5-vl} backbone, limiting its adaptability.
Moreover, \ourmodel{} delivers superior performance on most of these benchmarks when compared to specialized understanding models like Qwen2.5-VL and InternVL2.5~\cite{internvl2.5}, demonstrating that our MoT design effectively mitigates task conflicts while maintaining strong visual understanding capabilities.

\subsection{Image Generation}\label{subsec:vis_gen}
\begin{table*}[ht]
    \centering
    \setlength{\tabcolsep}{4pt}
    \renewcommand{\arraystretch}{1.2}
    \scriptsize
    \caption{\textbf{Evaluation of text-to-image generation ability on GenEval benchmark.} `Gen. Only' stands for an image generation model, and `Unified' denotes a model that has both understanding and generation capabilities.
    $\dagger$ refer to the methods using LLM rewriter.
    }
    \begin{tabular}{clccccccc}
        \toprule
        \textbf{Type} & \textbf{Model}  & \textbf{Single Obj.} & \textbf{Two Obj.} & \textbf{Counting} & \textbf{Colors} & \textbf{Position} & \textbf{Color Attri.} & \textbf{Overall$\uparrow$} \\
        \midrule
        \multirow{8}{*}{\rotatebox{90}{\textit{Gen. Only}}}
        & PixArt-$\alpha$~\cite{chen2024pixart} &  0.98 & 0.50 & 0.44 & 0.80 & 0.08 & 0.07 & 0.48 \\
        & SDv$2.1$~\cite{rombach2022high} & 0.98 & 0.51 & 0.44 & 0.85 & 0.07 & 0.17 & 0.50 \\
        & DALL-E $2$~\cite{dalle2}  & 0.94 & 0.66 & 0.49 & 0.77 & 0.10 & 0.19 & 0.52 \\
        & Emu$3$-Gen ~\cite{emu3}  & 0.98 & 0.71 & 0.34 & 0.81 & 0.17 & 0.21 & 0.54 \\
        & SDXL~\cite{sdxl} &  0.98 & 0.74 & 0.39 & 0.85 & 0.15 & 0.23 & 0.55 \\
        & DALL-E $3$~\cite{dalle3} & 0.96 & 0.87 & 0.47 & 0.83 & 0.43 & 0.45 & 0.67 \\
        & SD3-Medium~\cite{SD3} & 0.99 & 0.94 & 0.72 & 0.89 & 0.33 & 0.60 & 0.74 \\
        & FLUX.1-dev$^{\dagger}$~\cite{flux} & 0.98 & 0.93 & 0.75 & 0.93 & 0.68 & 0.65 & \emph{0.82} \\
        \midrule
        \multirow{13}{*}{\rotatebox{90}{\textit{Unified}}}
        & Chameleon~\cite{chameleon} &  - & - & - & - & - & - & 0.39 \\
        & LWM~\cite{lwm} &  0.93 & 0.41 & 0.46 & 0.79 & 0.09 & 0.15 & 0.47 \\
        & SEED-X~\cite{seed-x}  & 0.97 & 0.58 & 0.26 & 0.80 & 0.19 & 0.14 & 0.49 \\
        & TokenFlow-XL~\cite{qu2024tokenflow} &  0.95 & 0.60 & 0.41 & 0.81 & 0.16 & 0.24 & 0.55 \\
        & ILLUME~\cite{wang2024illume} &  0.99 & 0.86 & 0.45 & 0.71 & 0.39 & 0.28 & 0.61 \\
        & Janus~\cite{janus2024} & 0.97 & 0.68 & 0.30 & 0.84 & 0.46 & 0.42 & 0.61 \\
        & Transfusion~\cite{transfusion} & - & - & - & - & - & - & 0.63 \\
        & Emu$3$-Gen$^{\dagger}$\cite{emu3} & 0.99 & 0.81 & 0.42 & 0.80 & 0.49 & 0.45 & 0.66 \\
        & Show-o~\cite{show-o} &  0.98 & 0.80 & 0.66 & 0.84 & 0.31 & 0.50 & 0.68 \\
        & Janus-Pro-7B~\cite{januspro2025} &  0.99 & 0.89 & 0.59 & 0.90 & 0.79 & 0.66 & 0.80 \\
        & MetaQuery-XL$^{\dagger}$~\cite{pan2025transfer} &  -& - & - & -& -& -& 0.80 \\
        \rowcolor{myblue}
        & \textbf{BAGEL} & 0.99 & 0.94  & 0.81 & 0.88 & 0.64 &0.63 & \emph{0.82}\\
    \rowcolor{myblue}
    & \textbf{BAGEL}$^{\dagger}$ & 0.98 & 0.95  & 0.84 & 0.95 & 0.78 &0.77 & \textbf{0.88} \\
    \bottomrule
    \end{tabular}
    \label{tab:geneval}
\end{table*}

\begin{table*}[!ht]
    \centering
    \scriptsize
    \caption{\textbf{Comparison of world knowledge reasoning on WISE.} WISE examines the complex semantic understanding and world knowledge for T2I generation. `Gen. Only' stands for an image generation model, and `Unified' denotes a model that has both understanding and generation capabilities.
    **: Results of GPT-4o are tested by \cite{gpt_image_eval}.
    }
\label{tab:wisescore}
    \setlength{\tabcolsep}{8pt}
    \begin{tabular}{clccccccc}
    \toprule
    \textbf{Type} & \textbf{Model} & \textbf{Cultural}  & \textbf{Time}     & \textbf{Space}    & \textbf{Biology}    & \textbf{Physics} & \textbf{Chemistry} & \textbf{Overall$\uparrow$} \\
    \midrule
            \multirow{6}{*}{\rotatebox{90}{\textit{Gen. Only}}} &
 SDv1.5~\cite{rombach2022high} & 0.34 & 0.35& 0.32&0.28 &0.29 &0.21 & 0.32\\
& SDXL~\cite{sdxl} &0.43  & 0.48 &0.47  &0.44  &0.45 &0.27 & 0.43 \\
& SD3.5-large~\cite{SD3} & 0.44 &0.50 &0.58  & 0.44&0.52 &0.31 & 0.46 \\
& PixArt-Alpha~\cite{chen2024pixart} & 0.45  & 0.50& 0.48 & 0.49& 0.56 &0.34 & 0.47\\
& playground-v2.5~\cite{li2024playground} & 0.49  &0.58  & 0.55&0.43  & 0.48&0.33 & 0.49 \\
& FLUX.1-dev~\cite{flux} & 0.48  &0.58 &0.62  &0.42  &0.51 & 0.35 & 0.50 \\
    \midrule
    \multirow{9}{*}{\rotatebox{90}{\textit{Unified}}} 
& Janus~\cite{janus2024} &0.16 &0.26 &0.35 & 0.28 &0.30 & 0.14& 0.23\\
 &VILA-U~\cite{vila-u} & 0.26 &0.33  & 0.37 &0.35  &0.39 &0.23 & 0.31\\
& Show-o-512~\cite{show-o} & 0.28 &0.40  &0.48 & 0.30& 0.46 & 0.30 & 0.35\\
& Janus-Pro-7B~\cite{januspro2025} & 0.30& 0.37& 0.49 & 0.36&0.42 &0.26 & 0.35 \\
& Emu3~\cite{emu3} & 0.34 & 0.45 & 0.48 & 0.41  & 0.45 & 0.27 & 0.39 \\
&MetaQuery-XL~\cite{pan2025transfer} & 0.56& 0.55 &0.62 &  0.49 &  0.63 & 0.41 & 0.55 \\
&GPT-4o$**$ &0.81 &0.71 &0.89 &0.83 &0.79 &0.74 & \textbf{0.80} \\
\rowcolor{myblue}
& \textbf{BAGEL} & 0.44 & 0.55 & 0.68 & 0.44 & 0.60 & 0.39 & 0.52 \\
\rowcolor{myblue}
& \textbf{BAGEL} $w/$ Self-CoT & 0.76 & 0.69 & 0.75 & 0.65 & 0.75 & 0.58 & \emph{0.70} \\
        \bottomrule
    \end{tabular}
\end{table*}

We evaluate visual generation performance on two benchmarks: GenEval and WISE. As shown in \Cref{tab:geneval}, under the same evaluation settings as MetaQuery-XL, \ourmodel{} achieves an \textbf{88\%} overall score, outperforming both specialized generation models (FLUX-1-dev: 82\%, SD3-Medium: 74\%) and unified models (Janus-Pro: 80\%, MetaQuery-XL: 80\%). Even without an LLM rewriter, \ourmodel{} attains 82\%, surpassing the previous SOTA unified model, Janus-Pro-7B. 
On the WISE benchmark, \ourmodel{} exceeds all prior models except the leading private model, GPT-4o. It indicates that \ourmodel{} has strong reasoning ability with world knowledge.

\begin{figure*}[!htbp]
  \vspace*{-0.14\textwidth}
  \hspace*{-0.05\textwidth}
  \includegraphics[width=1.1\textwidth]{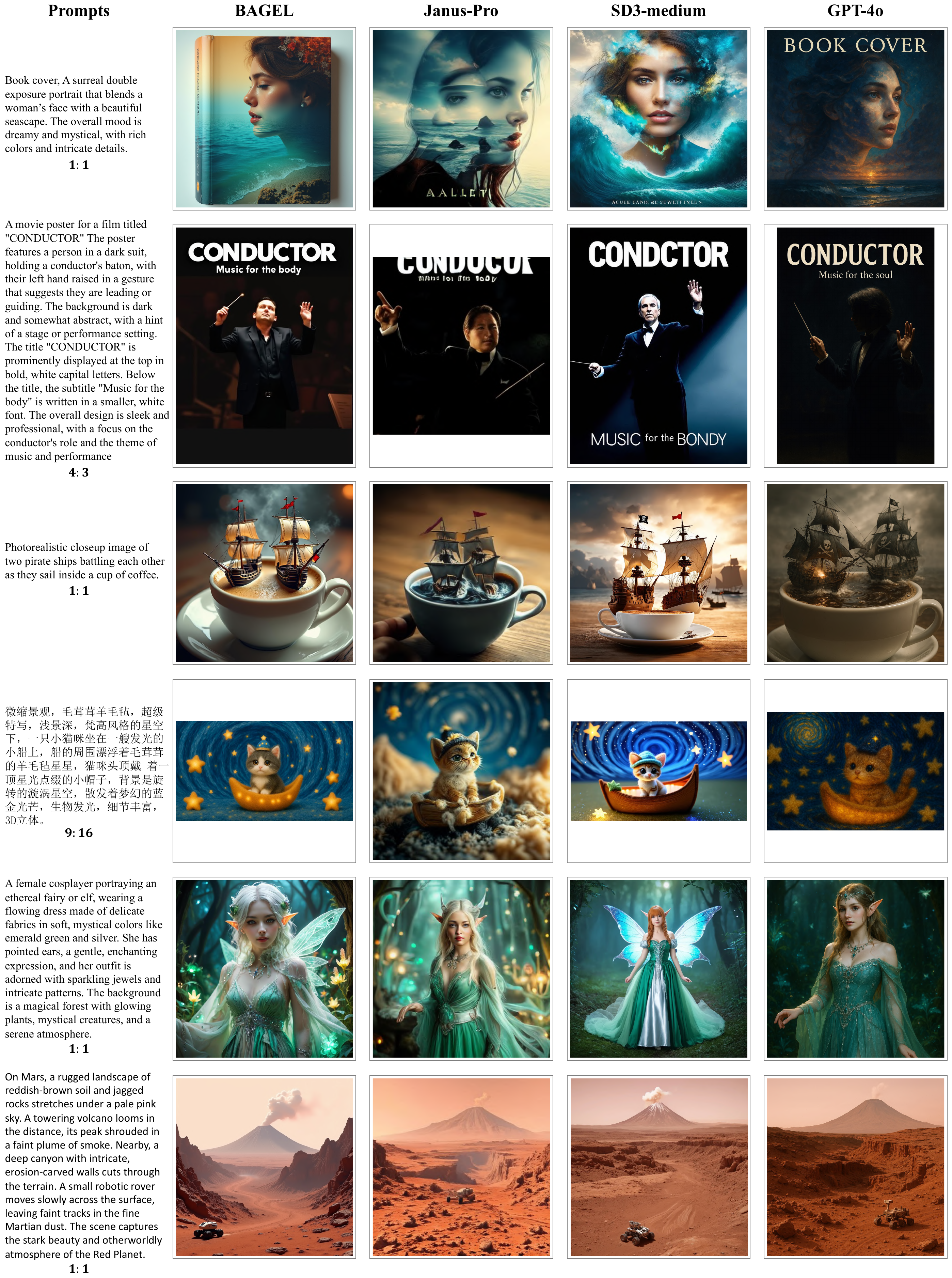}
  \caption{\textbf{T2I qualitative comparison.} Note that SD3-medium cannot take Chinese prompts so we translate them into English. For GPT-4o, we control the aspect ratio via text prompt. JanusPro can only generates square images.}
    \label{fig:t2i}
\end{figure*}

We conduct a qualitative comparison between \ourmodel{} and Janus‑Pro 7B, SD3‑medium, and GPT‑4o. As shown in \Cref{fig:t2i}, \ourmodel{} generates significantly higher‑quality images than Janus‑Pro~7B and also surpasses the widely used specialist text‑to‑image model SD3‑medium. Moreover, it natively supports prompts in both Chinese and English and allows generation at arbitrary aspect ratios.

\subsection{Image Editing}\label{subsec:editing}

\begin{table*}[ht]
    \centering
    \scriptsize
\begin{tabular}{cl|ccc|ccc}
\toprule
\multirow{2}{*}{\textbf{Type}} & \multirow{2}{*}{\textbf{Model}} &
\multicolumn{3}{c|}{\textbf{GEdit-Bench-EN (Full set)$\uparrow$}} 
& \multicolumn{3}{c}{\textbf{GEdit-Bench-CN (Full set)$\uparrow$}} \\
\cmidrule(lr){3-8}
& & \textbf{G\_SC} & \textbf{G\_PQ} & \textbf{G\_O} 
& \textbf{G\_SC} & \textbf{G\_PQ} & \textbf{G\_O} \\
\midrule
\multirow{2}{*}{\textit{Private}}&Gemini 2.0~\cite{gemini220250312}                       & 6.73 & 6.61 & 6.32 & 5.43 & 6.78 & 5.36  \\
&GPT-4o~\cite{openai2025chatgpt4o}                         & \textbf{7.85} & \textbf{7.62} & \textbf{7.53} & \textbf{7.67} & \textbf{7.56} & \textbf{7.30}  \\
\midrule
\multirow{6}{*}{\textit{Open-source}}
&Instruct-Pix2Pix~\cite{Brooks2022InstructPix2PixLT} & 3.58 & 5.49 & 3.68 & - & - & - \\
&MagicBrush~\cite{zhang2023magicbrush}               & 4.68 & 5.66 & 4.52 &  - & - & - \\
&AnyEdit~\cite{yu2024anyedit}                        & 3.18 & 5.82 & 3.21 &  - & - & - \\
&OmniGen~\cite{xiao2024omnigen}                      & 5.96 & 5.89 & 5.06 & - & - & - \\
&Step1X-Edit~\cite{liu2025step1xeditpracticalframeworkgeneral}& 7.09 & 6.76 & \emph{6.70} & 7.20 & \emph{6.87} &\emph{6.86}  \\
\rowcolor{myblue}
& \textbf{BAGEL}  &\emph{7.36} & \emph{6.83}& 6.52 & \emph{7.34} & 6.85 & 6.50 \\
\bottomrule

\end{tabular}
\caption{\textbf{Comparison on GEdit-Bench.} All metrics are reported as higher-is-better ($\uparrow$). G\_SC, G\_PQ, and G\_O refer to the metrics evaluated by GPT-4.1.}
\label{tab:gedit}
\vspace{-1pt}
\end{table*}

\begin{table}[!h]
    \centering
    \scriptsize
    \caption{\textbf{Comparison on IntelligentBench.} IntelligentBench examines complex reasoning ability in an image-editing context. $**$: Results are reported only on the subset of cases answered (some responses were rejected). GPT-4o answered 318 of 350 questions, while Gemini 2.0 answered 349 questions.}
\label{tab:IntelligentBench}
    \setlength{\tabcolsep}{8pt}
    \begin{tabular}{clc}
    \toprule
    \textbf{Type} & \textbf{Model}  & \textbf{Score$\uparrow$} \\
    \midrule
    \multirow{2}{*}{\textit{Private}}
       & GPT-4o$**$~\cite{openai2025chatgpt4o} & 78.9\\
       & Gemini 2.0$**$~\cite{gemini220250312} & 57.6 \\
    \midrule
        \multirow{3}{*}{\textit{Open-source}}& Step1X-Edit~\cite{liu2025step1xeditpracticalframeworkgeneral} & 14.9\\ 
& \textbf{BAGEL} & 44.9 \\
& \textbf{BAGEL} $w/$ Self-CoT & 55.3 \\
        \bottomrule
    \end{tabular}

\end{table}

We further evaluate the classical image editing capabilities of \ourmodel{} using the GEdit-Bench~\cite{liu2025step1x}. As shown in \Cref{tab:gedit}, \ourmodel{} achieves results competitive with the current leading specialist image editing model Step1X-Edit~\cite{liu2025step1x}, and also outperforms Gemini 2.0. Additionally, we report results on our newly proposed IntelligentBench in \Cref{tab:IntelligentBench}, where \ourmodel{} attains a performance of $44.9$, significantly surpassing the existing open-source Step1X-Edit model by $30$.

\begin{figure*}[!htbp]     
  \vspace*{-0.1\textwidth}
  \hspace*{-0.1\textwidth}
  \includegraphics[width=1.15\textwidth]{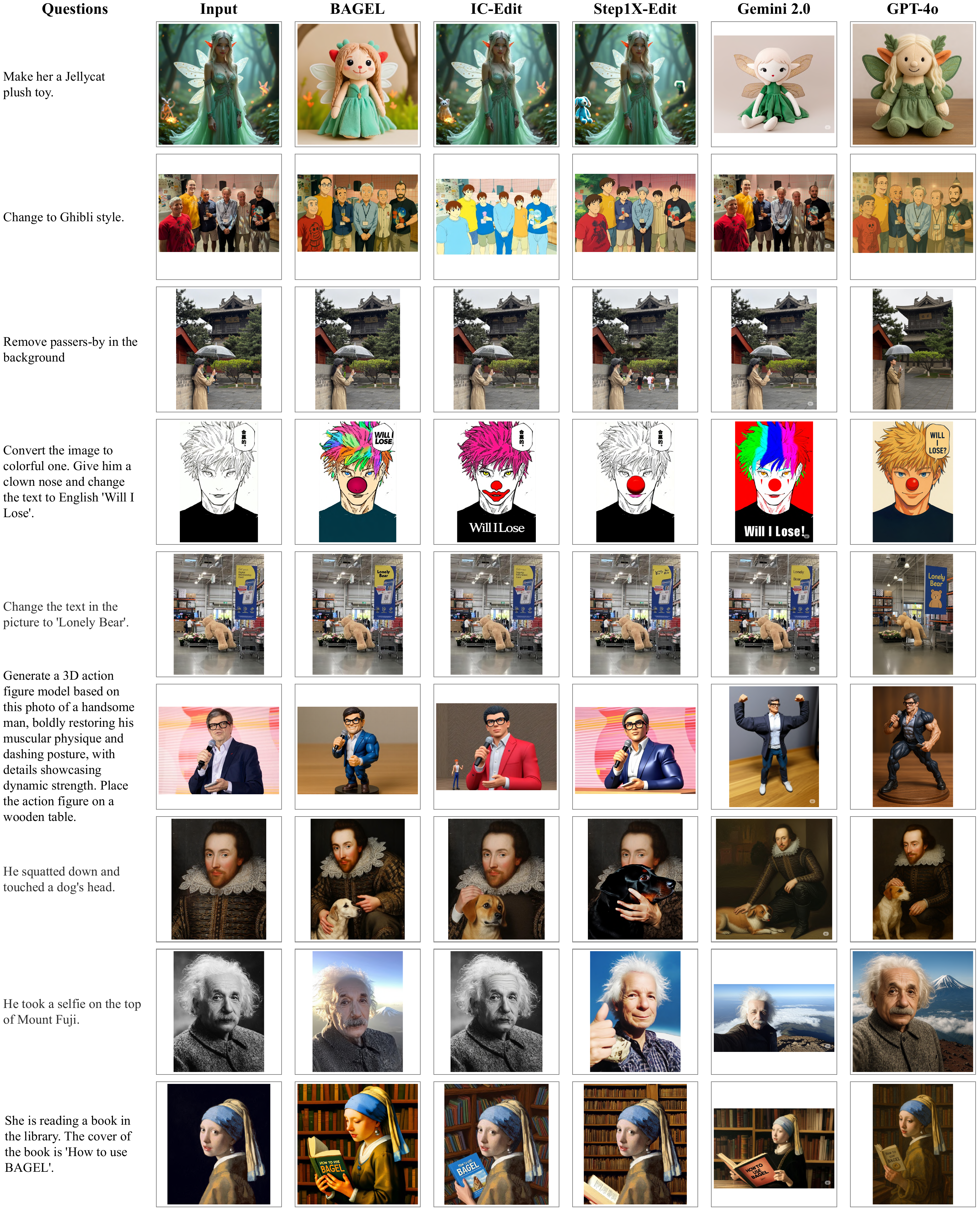}
  \caption{\textbf{Comparison on editing and manipulation tasks.}}
\label{fig:normal_edit}
\end{figure*}

\begin{figure*}[!htbp]     
  \vspace*{-0.1\textwidth}
  \hspace*{-0.1\textwidth}
  \includegraphics[width=1.15\textwidth]{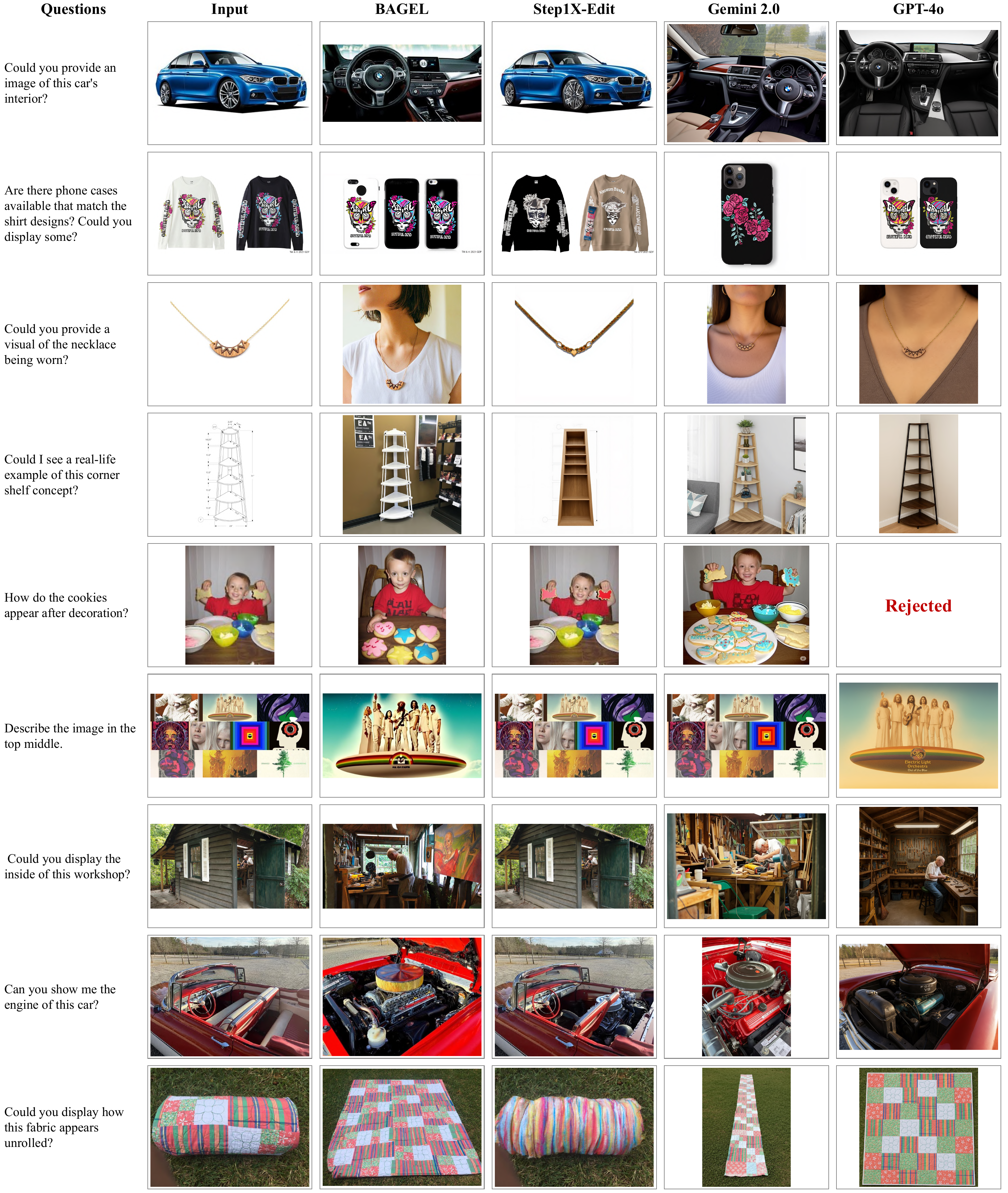}
  \caption{\textbf{Comparison on IntelligentBench.} The results demonstrate that (i) BAGEL achieves performance comparable to Gemini 2.0, effectively handling complex cases that require multi-step reasoning and the incorporation of world knowledge; and (ii) Step1X-Edit fails to address certain instances, often producing outputs that closely resemble the input image, which may be attributed to its limited reasoning capabilities. Note that BAGEL results here are generated in thinking mode.}
\label{fig:Intelligent_all}
\end{figure*}

We also provide qualitative comparisons across a diverse set of image editing scenarios in \Cref{fig:normal_edit} and \Cref{fig:Intelligent_all}, benchmarking \ourmodel{} against Gemini 2.0, GPT-4o, Step1X-Edit, and IC-Edit~\cite{zhang2025ICEdit}. As illustrated, \ourmodel{} consistently demonstrates superior performance over Step1X-Edit and IC-Edit, and also exceeds the capabilities of Gemini 2.0. While GPT-4o successfully handles these scenarios, it tends to introduce unintended modifications to the source images, an issue that \ourmodel{} effectively avoids.

\subsection{Generation/Editing with Thinking}\label{subsec:thing_gen}
In this section, we validate the effectiveness of reasoning-augmented generation across various benchmarks from both quantitative and qualitative perspectives.

\vspace{5pt}

\begin{figure*}[!htbp]
\vspace{-0.12\textwidth}
\centering
\begin{subfigure}[b]{1.0\textwidth}
    \centering
    \includegraphics[width=\textwidth]{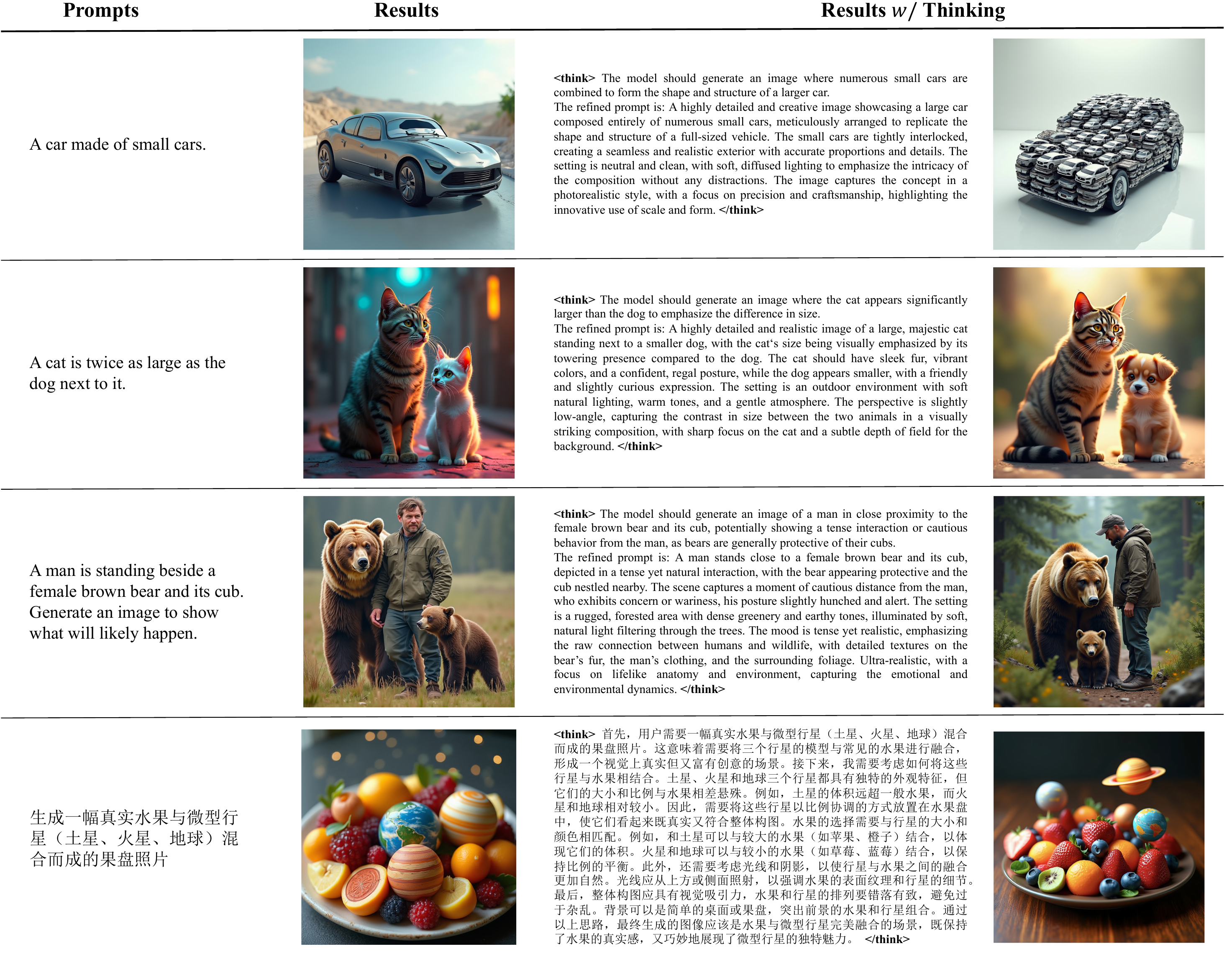}
    \vspace{-1.5em}
    \caption{\scriptsize{Thinking Helps Generation: Text-to-Image Generation Cases}}
    \label{fig:think_help_gen_t2i}
\end{subfigure}

\vspace{0.5em}

\begin{subfigure}[b]{1.0\textwidth}
    \centering
    \includegraphics[width=\textwidth]{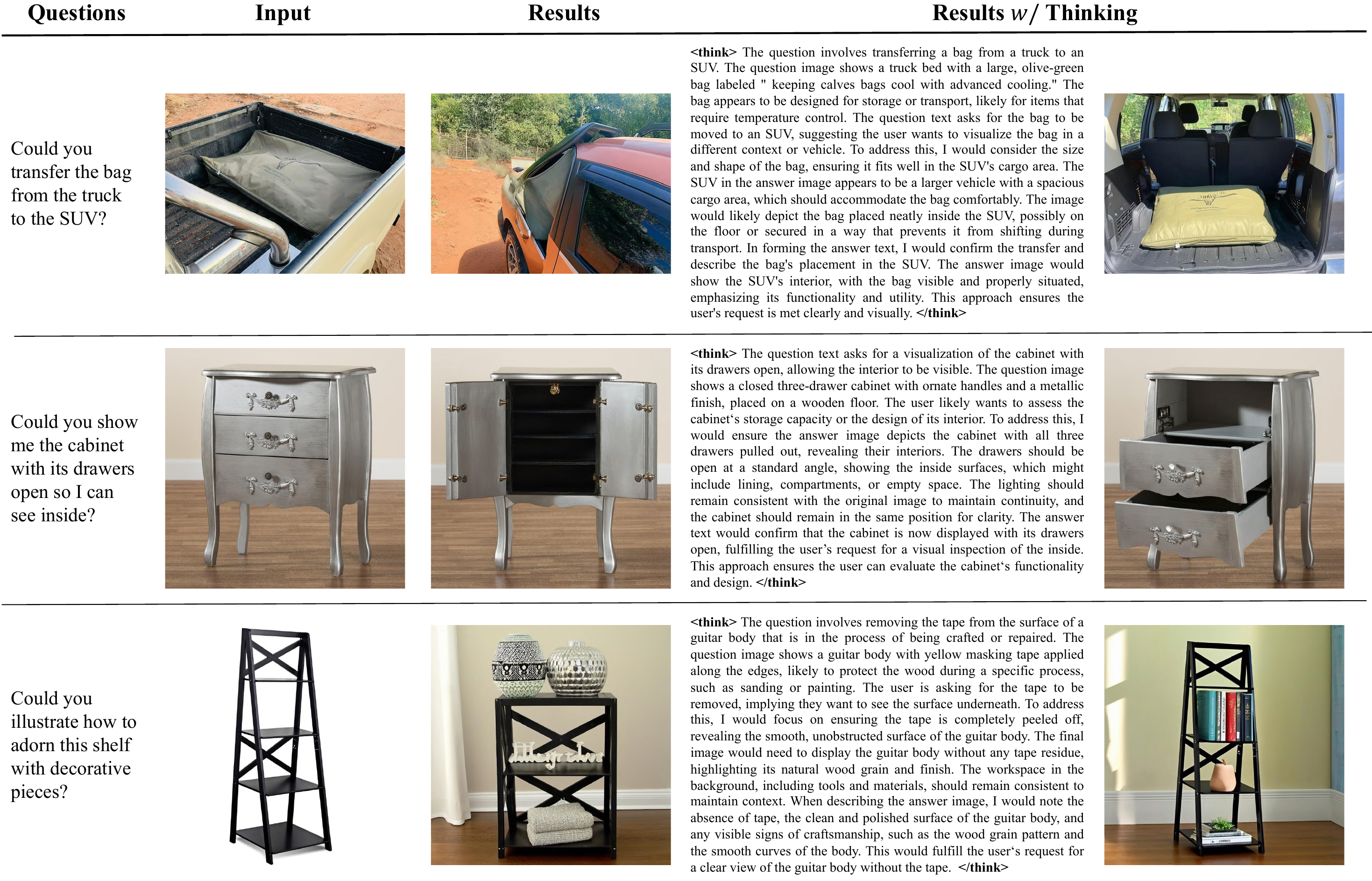}
    \vspace{-1.5em}
    \caption{\scriptsize{\textbf{Thinking Helps Generation: Image Editing Cases}}}
    \label{fig:think_help_gen_edit}
\end{subfigure}
\vspace{-1.5em}
\caption{Illustration of thinking-aided generation in two tasks. (a) Text-to-image generation. (b) Intelligent editing.}
\label{fig:think_help_gen_combined}
\end{figure*}

\textbf{Generation with thinking.} For Text-to-Image task, we evaluate Bagel on WISE with explicit chain-of-thought (CoT) reasoning process before generation. As shown in \Cref{tab:wisescore}, BAGEL with CoT achieves a score of $0.70$, surpassing its non-CoT counterpart by $0.18$, and also outperforms all existing open-source models by a significant margin (previous SOTA: MetaQuery-XL at $0.55$). In addition to the quantitative evaluation, we provide visualizations in \Cref{fig:think_help_gen_t2i}, where BAGEL fails to generate correct images when given only a short prompt, but succeeds when using the CoT-based thinking paradigm.

\textbf{Editing with Thinking.}
As presented in \Cref{tab:IntelligentBench}, incorporating CoT into BAGEL improves its Intelligent Score from $44.9$ to $55.3$.
This performance gain is primarily attributed to the inclusion of reasoning, which enables the model to leverage world knowledge and provide detailed editing guidance. 
Consistent improvements are also observed on RISEBench~\cite{rise} (\Cref{tab:rise}, from $6.1$ to $11.9$) and KRIS-Bench~\cite{kris} (\Cref{tab:kris}, from $56.21$ to $60.18$).
We further illustrate several representative cases from IntelligentBench in \Cref{fig:think_help_gen_edit}, where the tasks demand general knowledge or multi-step reasoning. In these scenarios, BAGEL demonstrates significantly improved image editing capabilities when guided by the thinking content.

\subsection{World Modeling}\label{subsec:worldnav}
\begin{figure*}[!htbp]
  \hspace*{-0.11\textwidth}
  \includegraphics[width=1.2\textwidth]{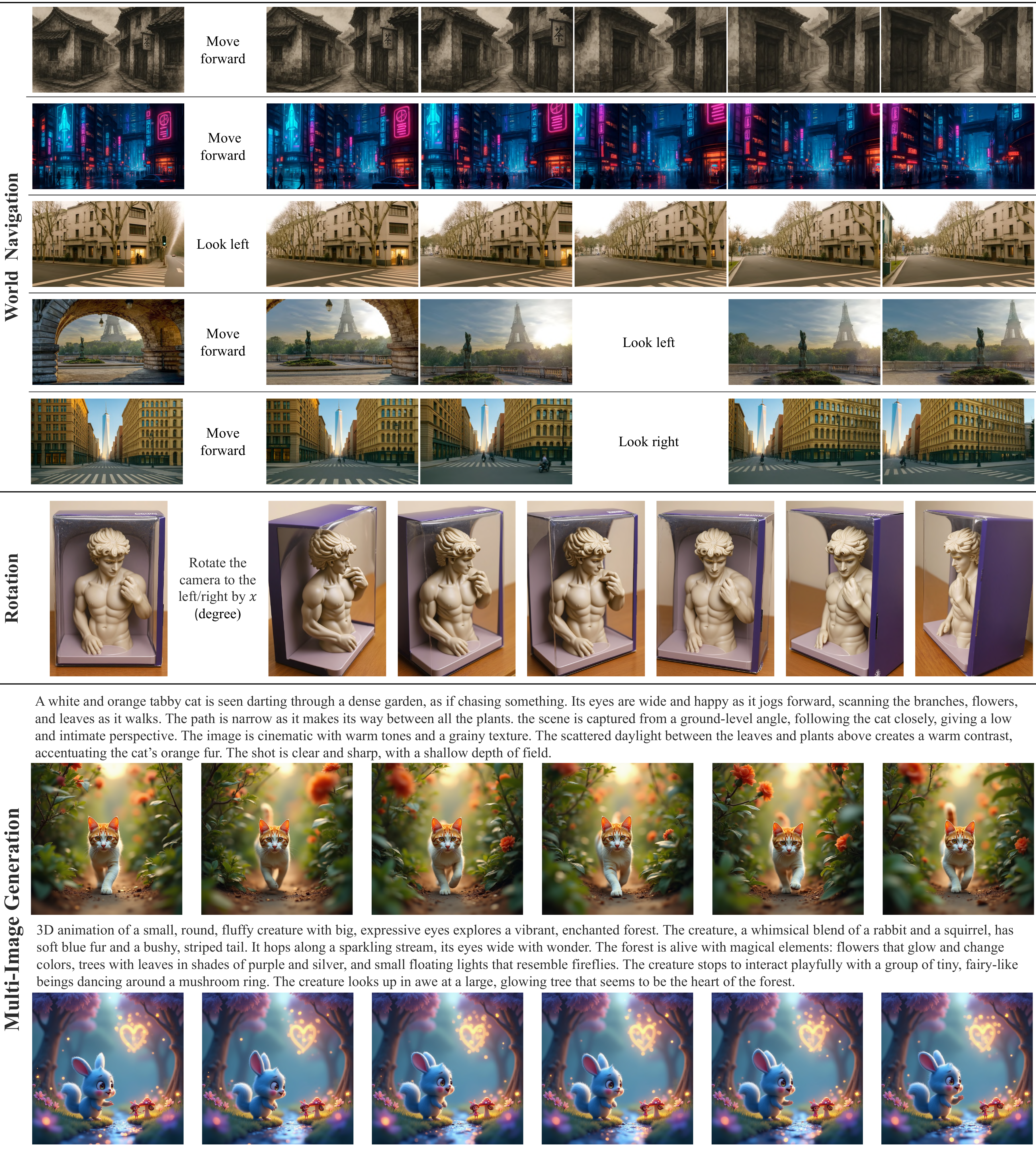}
  \caption{\textbf{Examples of \ourmodel{} in navigation, rotation, and multi-image generation.}}
  \label{fig:worldmodel}
\end{figure*}

To improve BAGEL's world modeling ability for long-sequence visual generation, we fine-tune the model by increasing the proportion of video and navigation data in the training recipe.
For navigation, we construct our dataset from video interleave sequences, annotating camera trajectories using ParticleSfM~\cite{zhao2022particlesfm}. 
In \Cref{fig:worldmodel}, we demonstrate \ourmodel{}'s world modeling capabilities, which include world navigation, rotation, and multi-frame generation. 

From the figure, \ourmodel{} exhibits robust world understanding and simulation capabilities. It can follow input instructions to generate a dynamic number of images for tasks like navigating and rotating an input image, or produce multiple images based on a given prompt. Additionally, \ourmodel{} demonstrates strong generalization in world understanding. For instance, while trained solely on real-world street navigation, it seamlessly extends to diverse domains such as ink paintings, cartoons, and video games.

\subsection{More Qualitative Results}

\begin{figure*}[!htbp]     
  \vspace*{-0.1\textwidth}
  \includegraphics[width=1.0\textwidth]{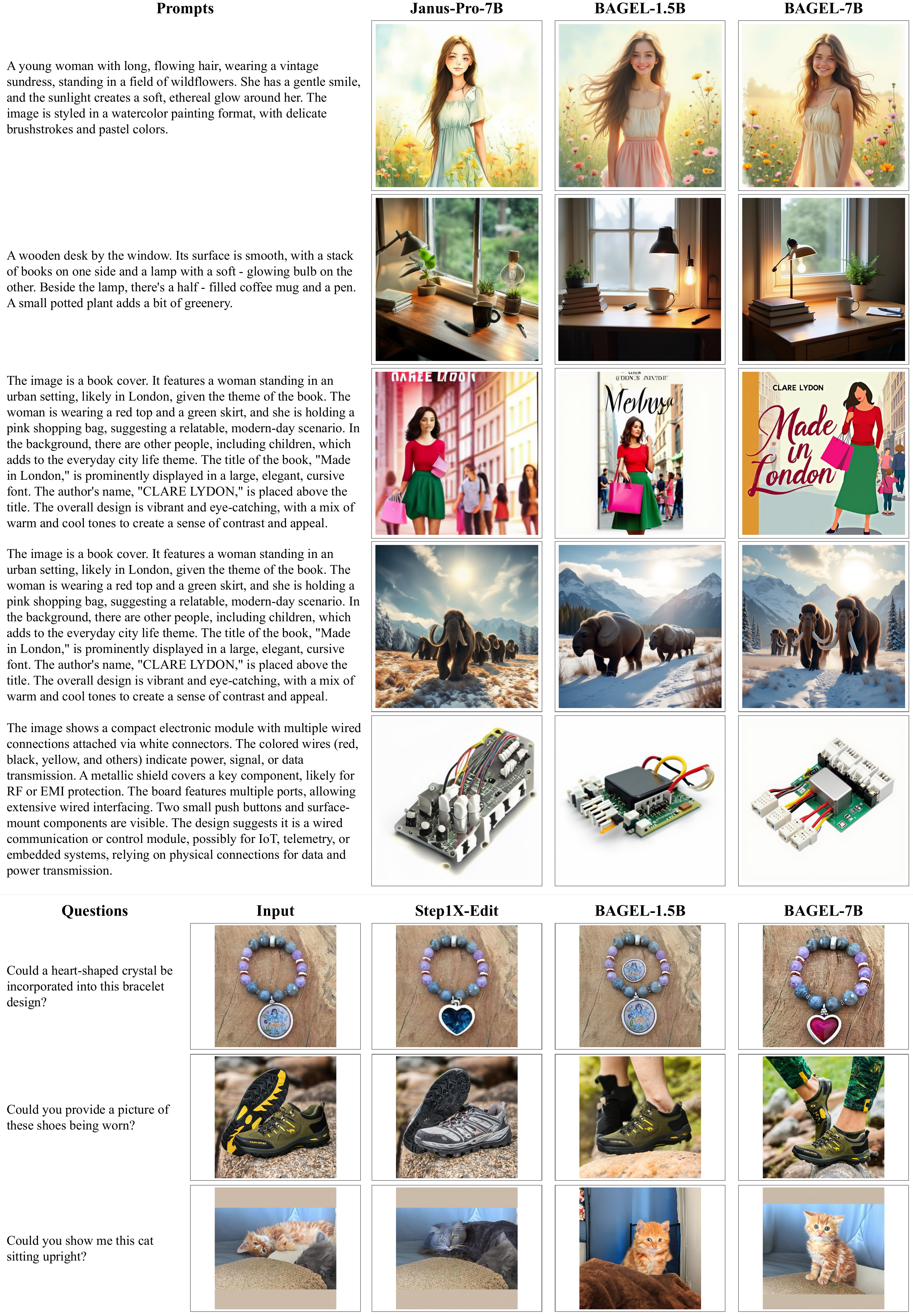}
  \caption{\textbf{Effect of model scaling:} larger models demonstrate better prompt adherence and produce higher-quality images.}
\label{fig:bagel_compare}
\end{figure*}

\textbf{Performance of {BAGEL-1.5B}.} \Cref{fig:bagel_compare} compares the text-to-image (T2I) and image-editing performance of {BAGEL-1.5B}—with 1.5 B activated parameters—against JanusPro-7B and Step1X-Edit (12B). Although {BAGEL-1.5B} is considerably smaller, it surpasses both larger models on both tasks in terms of qualitative comparison. Moreover, the gap between {BAGEL-1.5B} and {BAGEL-7B} underscores the gains from model scaling, indicating a greater potential for even larger BAGEL variants.

\begin{figure*}[!htbp]     
  \vspace*{-0.12\textwidth}
  \hspace*{-0.1\textwidth}
  \includegraphics[width=1.175\textwidth]{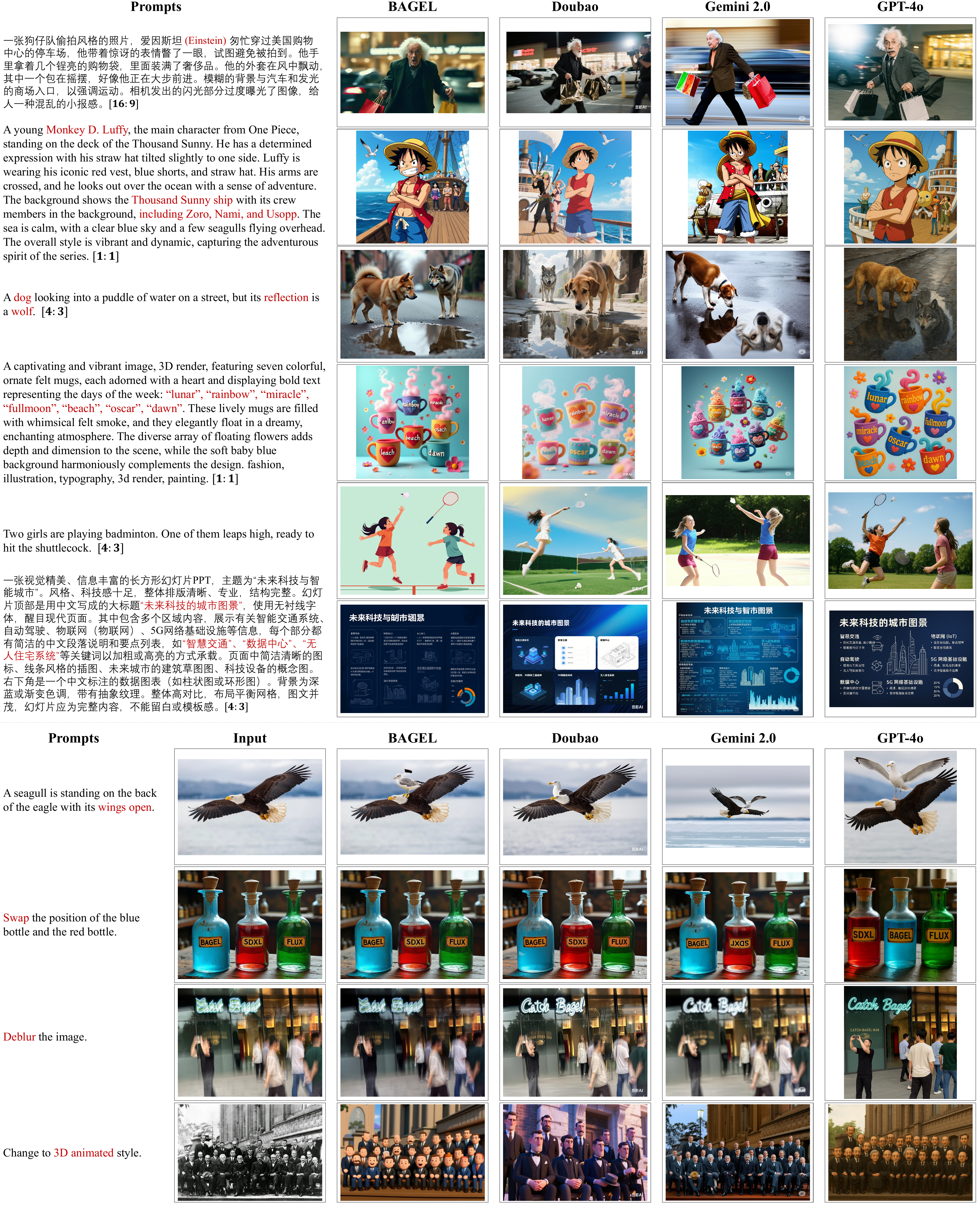}
  \caption{\textbf{Failure cases.} Tasks involving certain IP, complicated text, counterfactual scenes, object swapping, and deblurring pose challenges for BAGEL and other models. In contrast, GPT-4o demonstrates more consistent success in these scenarios.
  }
\label{fig:failure}
\end{figure*}

\textbf{Failure cases.}
In \Cref{fig:failure} we present representative failure cases for BAGEL alongside other state-of-the-art models. Tasks that feature special IP generation, complex textual rendering, intricate human pose generation, or the simultaneous generation of multiple instances remain persistently challenging for contemporary text-to-image systems.
For image editing, operations such as swapping object positions or simultaneously modifying a large amount of instances likewise challenge most existing models. In some complex scenarios, both BAGEL and Gemini 2.0 exhibit similar difficulties in adhering precisely to the given instructions. By contrast, {GPT-4o} delivers the most consistently successful results across all examples. 
Performance of BAGEL can be simply enhanced by scaling data with additional text-containing images, increasing model capacity, or applying RLHF~\cite{rlhf} during the final post-training stage.
\section{Conclusion}

We presented \ourmodel{}, a unified multimodal understanding and generation model that shows emerging capabilities when scaling up unified pretraining. \ourmodel{} yields top-tier performance on standard multimodal understanding and generation benchmarks, and further distinguish itself with powerful world modeling and reasoning capabilities. In the hope of unlocking further opportunities for multimodal research, we open source BAGEL to the research community.

\section{Acknowledgement}
We'd like to thank Ziqian Wei, Haoli Chen, Shengyang Xu, Chen Li, Yujia Qin, Yi Lin, Wenhao Huang, Shen Yan, Xiaojun Xiao, Yan Wu, Gang Wu, Guodong Li, Kang Lei, Liang-Wei Tao, Qifan Yang, Bairen Yi, Xiuli Chen, Rui Cao, Yating Wang, Yufeng Zhou, Mingdi Xu, Tingting Zhang, Xuehan Xiong, Tianheng Cheng, Zanbo Wang, Heng Zhang, Yanghua Peng, Faming Wu, Jiashi Feng, Jianfeng Zhang, Xiu Li for their contributions to the BAGEL project.

\clearpage

\bibliographystyle{plainnat}
\bibliography{main}

\clearpage

\beginappendix

\begin{table}[ht]
    \centering
    \scriptsize
\begin{tabular}{cl|ccccc}
\toprule
\textbf{Type} & \textbf{Model} &
\textbf{Temporal} & \textbf{Causal} & \textbf{Spatial} & \textbf{Logical} & \textbf{Overall$\uparrow$} \\
\midrule
\multirow{2}{*}{\textit{Private}}&Gemini 2.0~\cite{gemini220250312}                       & 8.2 & 15.5 & 23.0 & 4.7 & 13.3 \\
&GPT-4o~\cite{openai2025chatgpt4o} & 34.1 & 32.2 & 37.0 & 10.6 & 28.9  \\
\midrule
\multirow{5}{*}{\textit{Open-source}} & EMU2~\cite{emu2} & 1.2 & 1.1 & 0.0 & 0.0 & 0.5 \\
&OmniGen~\cite{xiao2024omnigen} & 1.2 & 1.0 & 0.0 & 1.2 & 0.8 \\
&Step1X-Edit~\cite{liu2025step1xeditpracticalframeworkgeneral}& 0.0 & 2.2 & 2.0 & 3.5 & 1.9 \\
\rowcolor{myblue}
& \textbf{BAGEL} & 2.4 & 5.6 & 14.0 & 1.2 & 6.1 \\
\rowcolor{myblue}
& \textbf{BAGEL} $w/$ Self-CoT & 5.9 & 17.8 & 21.0 & 1.2 & 11.9 \\
\bottomrule
\end{tabular}
\caption{\textbf{Comparison on RISEBench.} Results are evaluated by GPT-4.1.}
\label{tab:rise}
\vspace{-1pt}
\end{table}

\begin{table}[ht]
    \centering
    \scriptsize
\begin{tabular}{cl|ccc|cc|cc|c}
\toprule
\multirow{2}{*}{\textbf{Type}} & \multirow{2}{*}{\textbf{Model}} &
\multicolumn{3}{c|}{\textbf{Factual$\uparrow$}}  & \multicolumn{2}{c|}{\textbf{Conceptual$\uparrow$}}  & \multicolumn{2}{c|}{\textbf{Procedural$\uparrow$}} & \multirow{2}{*}{\textbf{Overall$\uparrow$}} \\
\cmidrule(lr){3-9}
& & \textbf{AP} & \textbf{SP} & \textbf{TP} 
& \textbf{SS} & \textbf{NS} & \textbf{LP} & \textbf{ID} \\
\midrule
\multirow{2}{*}{\textit{Private}}&Gemini 2.0~\cite{gemini220250312}                       & 66.33 & 63.33 & 63.92 & 68.19 & 56.94 & 54.13 & 71.67 & 62.41 \\
& GPT-4o~\cite{openai2025chatgpt4o} & 83.17 & 79.08 & 68.25 & 85.50 & 80.06 & 71.56 & 85.08 & 80.09 \\
\midrule
\multirow{5}{*}{\textit{Open-source}} & EMU2~\cite{emu2} & 51.50 & 48.83 & 22.17 & 34.69 & 38.44 & 24.81 & 45.00 & 39.70 \\
&OmniGen~\cite{xiao2024omnigen} & 37.92 & 28.25 & 21.83 & 30.63 & 27.19 & 11.94 & 35.83 & 28.85 \\
&Step1X-Edit~\cite{liu2025step1xeditpracticalframeworkgeneral}& 55.50 & 51.75 & 0.00 & 44.69 & 49.06 & 40.88 & 22.75 & 43.29 \\
\rowcolor{myblue}
& \textbf{BAGEL} & 64.27 & 62.42 & 42.45 & 55.40 & 56.01 & 52.54 & 50.56 & 56.21 \\
\rowcolor{myblue}
& \textbf{BAGEL} $w/$ Self-CoT & 67.42 & 68.33 & 58.67 & 63.55 & 61.40 & 48.12 & 50.22 & 60.18 \\
\bottomrule
\end{tabular}
\caption{\textbf{Comparison on KRIS-Bench.} `AP', `SP', `TP', `SS', `NS', `LP' and `ID' represent `Attribute Perception', `Spatial Perception', `Temporal Prediction', `Social Science', `Natural Science', `Logical Reasoning' and `Instruction Decomposition', respectively. We report only the average metrics evaluated by GPT-4o.}
\label{tab:kris}
\vspace{-1pt}
\end{table}

\begin{table*}[!ht]\centering
\begin{minipage}{0.9\textwidth}\vspace{0mm}    
    \centering
    \begin{tcolorbox} 
        \centering
        \hspace{-6mm}
        \begin{tabular}{p{0.9\textwidth}}
        \hspace{1mm}
        \begin{minipage}{0.9\textwidth}
        \texttt{\#\#\#[System Role Instruction]}\\
You have the following information:\\
1. question image: [Place or reference the question image here]\\
2. question text: [Place the text of the question here]\\
3. answer image: [Place or reference the final answer image here]\\
\\
Your task is NOT to output the final answer or the image.  
Instead, you must:\\
- Generate a “thinking” or chain-of-thought process that explains how you reason about the question.\\
- Provide the reasoning/analysis that leads to the answer image.\\
- The reasoning/analysis should include what should be changed in the answer image compared to the question image and what should be kept the same.\\
- The reasoning should highlight that the input image structure and layout should be kept the same.\\
\\
Below is an example of how your output should look. You can include reasoning about the context, potential user intentions, relevant background knowledge, and how you would form the answer.  
The length of outputs should be **around or shorter than 60 tokens**. \\
\\
\textcolor{blue}{Example Output:}\\
The user wants to change the background from a sunny garden to a snowy setting. The structure and layout of the pink unicorn with bubble details and sunglasses should remain unchanged. Only the environment needs modification: replacing green grass with snow and surrounding greenery with frosted, snow-covered plants while maintaining lighting coherence.
        \end{minipage}
        \end{tabular}
    \end{tcolorbox}
    \vspace{-2mm}
    \caption{\textbf{The prompt to generate reasoning trace for Free-form image manipulation from edit data. }}
    \label{tab:prompt_edit}
    \end{minipage}
    \vspace{-2mm}
\end{table*}

\begin{table*}[!ht]\centering
\begin{minipage}{0.9\textwidth}\vspace{0mm}    
    \centering
    \begin{tcolorbox} 
        \centering
        \hspace{-6mm}
        \begin{tabular}{p{0.9\textwidth}}
        \hspace{1mm}
        \begin{minipage}{0.9\textwidth}
        \texttt{\#\#\#[System Role Instruction]}\\
You have the following information:\\
1. question image: [Place or reference the question image here]\\
2. question text: [Place the text of the question here]\\
3. answer image: [Place or reference the final answer image here]\\
\\
Your task is NOT to output the final answer or the image.  
Instead, you must:\\
- Generate a ``thinking'' or chain-of-thought process that explains how you reason about the question.\\
- Provide the reasoning/analysis that leads to the answer image.\\
- The reasoning/analysis should include what should be changed in the answer image compared to the question image and what should be kept the same.\\
\\
Below is an example of how your output should look. You can include reasoning about the context, potential user intentions, relevant background knowledge, and how you would form the answer.  
The length of outputs should be **around or shorter than 60 tokens**. \\
\\
\textcolor{blue}{Example Output:}
\\
First, I notice the cat's determined action in pressing a button. To adjust for the answer, the focus shifts to expressing excitement or eagerness. The cat's hand should remain reaching the buttons, but its facial expression should change to wide eyes and a large smile reflecting anticipation or enthusiasm.
        \end{minipage}
        \end{tabular}
    \end{tcolorbox}
    \vspace{-2mm}
    \caption{\textbf{The prompt to generate reasoning trace for Free-form image manipulation from video interleaved data. }}
    \label{tab:prompt_wmedit}
    \end{minipage}
    \vspace{-2mm}
\end{table*}

\begin{table*}[!ht]\centering
\begin{minipage}{0.9\textwidth}\vspace{0mm}    
    \centering
    \begin{tcolorbox} 
        \centering
        \hspace{-6mm}
        \begin{tabular}{p{0.9\textwidth}}
        \hspace{1mm}
        \begin{minipage}{0.9\textwidth}
        \texttt{\#\#\#[System Role Instruction]}\\
You have the following information: \\
1. question image: [Place or reference the question image here]\\
2. question text: [Place the text of the question here]\\
3. answer text: [Place the final answer text here]\\
4. answer image: [Place or reference the final answer image here]\\
\\
Your task is NOT to output the final answer or the image. Instead, you must:\\
- Generate a detailed “thinking” or chain-of-thought process that explains how you reason about the question.\\
- Do NOT include the final answer text in your output.\\
- Provide only the reasoning/analysis that leads to the final answer and the answer image (even though you will not reveal the final answer itself).\\
- The reasoning/analysis should include some description of the answer image to help the answer-image-generation.\\
\\
Below is an example of how your output should look.  
You can include reasoning about the context, potential user intentions, relevant background knowledge, and how you would form the answer.  
The length of outputs should be **around or shorter than 200 tokens**\\
\\
\textcolor{blue}{Example Output:}\\
First, I notice the user wants to see a vehicle displayed while it's moving. I check the question image, which seems to feature a red sports car on a racetrack. The question text, `Can you display the vehicle while it's moving?', suggests they want a visual depiction of a car in motion.  
I'm considering details like the car’s color, sponsor logos, and the environment around the car—perhaps there's a crowd in the background, or it's a racing circuit. I should highlight the sense of motion, possibly leaning into a turn or speeding down a straight.  
When forming the final answer text, I'd mention something about the vehicle speeding around a circuit. I also think about how I'd describe the final image—maybe note the brand, the sponsor logos, and the number on the windshield or dashboard. Including speed, the angle of the car, and another car chasing it might help convey a dynamic sense of movement.  
Lastly, I recall that the user specifically asked to `display the vehicle while it's moving,' so I'd ensure the image description references motion, leaning into a turn, and the impression of high velocity. This approach should fulfill their request.
        \end{minipage}
        \end{tabular}
    \end{tcolorbox}
    \vspace{-2mm}
    \caption{\textbf{The prompt to generate reasoning trace for conceptual editing.}}
    \label{tab:prompt_conceptedit}
    \end{minipage}
    \vspace{-2mm}
\end{table*}

\begin{table*}[!ht]\centering
\begin{minipage}{0.9\textwidth}\vspace{0mm}    
    \centering
    \begin{tcolorbox} 
        \centering
        \hspace{-6mm}
        \begin{tabular}{p{0.9\textwidth}}
        \hspace{1mm}
        \begin{minipage}{0.9\textwidth}
        \includegraphics[width=0.9\textwidth]{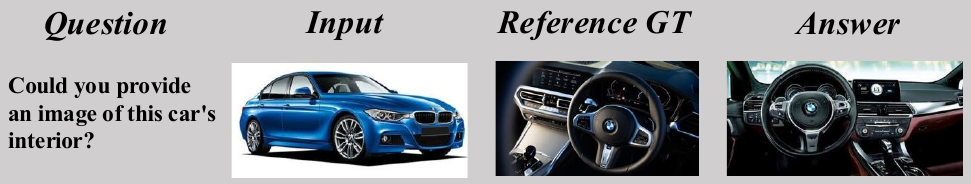}\\
        \texttt{\#\#\#Human:}\\
You are given a question, the corresponding question image, a human answered image, and the model-generated (AS) answer image. \\
Your task is to evaluate whether the AS answers the question based on the following criteria: \\
\textcolor{blue}{Must Exact Fulfillment of Request}: The answer image must fulfill the request made in the question. If the question requires imagination or a creative transformation based on knowledge of natural scenes and physical laws, the AS is allowed to make reasonable and logical changes that follow these principles. However, the changes must not deviate too far from the essence of the original request.

\textcolor{blue}{Must Satisfy Completeness}: Every element requested in the question must be reasonably present and completed in the answer image. Missing elements should be noted, but some degree of creative interpretation is acceptable as long as the request is overall fulfilled.

\textcolor{blue}{Must No Visual Errors}: The answer image must not contain major visual errors such as proportion issues, blurriness, or logical inconsistencies. Minor imperfections that do not affect the overall quality or coherence are acceptable, but significant visual errors should be avoided.

\textcolor{blue}{Can Allow Creative Changes Based on Knowledge}: If the question requires imaginative thinking or knowledge of natural scenes and physical laws, minor changes or additions that help fulfill the request are allowed. These changes should align with the natural world, physical principles, or the context of the question. However, large or inconsistent changes that break the scene's logic or introduce factual inaccuracies are not acceptable.

The human answered image is just an example answer for your reference to understand how to answer this question. The AS does not need to be the same as the human answered image.
\textcolor{blue}{You should assign a score based on how well the images meet these criteria}:

0: The AS can't be used for answering this question based on previous criteria. Compared with the AS, the human answered image is significantly better. \\
1: The AS can answer the question, but is worse than the human answered image in terms of quality.\\
2: The AS can answer the question with similar or better quality than the human answered image.\\
$\{$
  $"$score$":$ int, 
  $"$reason\_of\_score$":"$Detailed explanation of the reasoning for the score.$"$
$\}$

Now give me the accuracy score and reason strictly following the json format:
"""  \\\\
        \texttt{\#\#\#Assistant:} \\\textcolor{blue}{\{'reason'}: 'The AS correctly provides the interior image of a BMW car, which matches the request based on the exterior shown in the question image. The interior features, including the steering wheel with the BMW logo, dashboard, and overall design, are excellently captured without significant visual errors or missing elements. Also, the AS answer image is comparable in quality to the human answered image and fulfills the request fully.', \\
        \textcolor{blue}{'score'}: 2\}
        \end{minipage}
        \end{tabular}
    \end{tcolorbox}
    \vspace{-2mm}
    \caption{\textbf{Example of IntelligentBench Evaluation:} Please note that the "Human answered image" is simply a label for the GPT-4o's reference to the ground truth (GT) image, rather than an actual image generated by a human.}
    \label{tab:prompt_intelligentEditEval}
    \end{minipage}
    \vspace{-2mm}
\end{table*}

\end{document}